\pdfoutput=1
\documentclass[twocolumn]{styles/LaTeX_DL_468198/svjour3}  

\usepackage[utf8]{inputenc}
\usepackage{amsthm}
\usepackage{amsmath} 
\usepackage{amssymb}  
\usepackage{balance}

\usepackage{paralist}
\usepackage{todonotes}
\usepackage{hyperref}
\usepackage{csquotes}
\usepackage{balance}
\usepackage{amssymb}
\usepackage{algorithm}
\usepackage{algpseudocode}
\usepackage{varwidth}
\usepackage{booktabs} 
\usepackage{cases}
\usepackage{graphicx}
\usepackage{subcaption}
\usepackage[english]{babel}

\newcommand{\pname}  [1] {\textsc{#1}}

\makeatletter
\newlength{\limitedheadlinewidth}
\newlength{\limitedlinewidth}
\newtheoremstyle{qrock_definition}
  {6pt}
  {3pt}
  {
    \setlength{\limitedheadlinewidth}{\linewidth}
    \addtolength{\limitedheadlinewidth}{-.5em}
    \setlength{\limitedlinewidth}{\linewidth}
    \addtolength{\limitedlinewidth}{-1em}
    \parshape 2
            0.em  \limitedheadlinewidth
            0.5em \limitedlinewidth
  }
  {}
  {\bfseries}
  {.}
  {.5em}
  {\thmname{#1}\thmnumber{ #2}\thmnote{ (\textit{#3})}}

\theoremstyle{qrock_definition}
\newtheorem*{definition*}{Definition}	
\newtheorem*{assumption}{Assumption}

\renewcommand{\paragraph}{\@startsection{paragraph}{4}
    {0ex}
    {-3.25ex plus -1ex minus -0.2ex}%
    {1.5ex plus 0.2ex}
    {\normalfont\normalsize\itshape}}

\usepackage{xifthen}
\usepackage[final,xcolor=dvipdf]{changes}

\newcommand{\markChange}[3]{
    #3
}

\DeclareRobustCommand{\changed}[2]{
    \markChange{#1}{blue}{#2}
}

\usepackage{glossaries}
\glsdisablehyper
\newacronym{AI}{AI}{Artificial Intelligence}
\newacronym{AWS}{AWS}{Amazon Web Services}
\newacronym{AADL}{AADL}{Architecture Analysis and Design Language}
\newacronym{BM}{BM}{behavior model}
\newacronym{CAD}{CAD}{Computer Aided Design}
\newacronym[longplural={capabilities}]
    {CAP}{CAP}{capability}
\newacronym{CC}{CC}{cognitive core}
\newacronym{CF}{CF}{capability function}
\newacronym{CFM}{CFM}{capability function model}
\newacronym{CL}{Cluster}{cluster}
\newacronym{DB}{DB}{database}
\newacronym{DMP}{DMP}{Dynamic Movement Primitive}
\newacronym{ER}{ER}{Entity Relationship}
\newacronym{FF}{FF}{feature function}
\newacronym{FS}{FS}{feature space}
\newacronym{FT}{FT}{Feature}
\newacronym{GMM}{GMM}{Gaussian Mixture Model}
\newacronym{GUI}{GUI}{Graphical User Interface}
\newacronym{HyRoDyn}{HyRoDyn}{Hybrid Robot Dynamics}
\newacronym{Korcut}{Korcut}{Knowledge-based Open Robot voCabulary as Utility Toolkit}
\newacronym{KRR}{KR\&R}{Knowledge Representation and Reasoning}
\newacronym{HDDL}{HDDL}{Hierarchical Domain Definition Language}
\newacronym{HTN}{HTN}{Hierarchical Task Network}
\newacronym{ODE}{ODE}{ordinary differential equation}
\newacronym{PDDL}{PDDL}{Planning Domain Definition Language}
\newacronym{ROS}{ROS}{Robot Operating System}
\newacronym{Rock}{Rock}{Robot Construction Kit}
\newacronym{SA}{SA}{semantic annotation}
\newacronym{SDL}{SDL}{Semantic Description Language}
\newacronym{URDF}{URDF}{Universal Robot Description Format}
\newacronym{UML}{UML}{Unified Modelling Language}

\title{\LARGE \bf
    A Development Cycle for \\Automated Self-Exploration of Robot Behaviors
}

\author{Thomas M. Roehr$^{1,\dagger}$, Daniel Harnack$^{1}$, Hendrik W\"ohrle$^{1,3}$,\\
        Felix Wiebe$^{1}$, Moritz Schilling$^{^1}$, Oscar Lima$^{1}$, Malte Langosz$^{1}$, \\
        Shivesh Kumar$^{1}$, Sirko Straube$^{1}$, Frank Kirchner$^{1,2}$
    \thanks{$^{1}$DFKI GmbH Robotics Innovation Center, Bremen, Germany}%
    \thanks{$^{2}$AG Robotics, Department of Mathematics and Computer Science, University of Bremen, Germany}%
    \thanks{$^{3}$Institute for Communication Technology, Department of Information Technology, Dortmund University of Applied Sciences and Arts, Germany}%
    \thanks{$^{\dagger}$Corresponding author: {\tt\small thomas.roehr@dfki.de}}%
}

\makeglossaries

\begin{document}
    
    \maketitle
    \thispagestyle{empty}
    \pagestyle{empty}

    \begin{abstract}
        %
        In this paper we introduce \pname{Q-Rock}, 
        a development cycle for the automated self-exploration 
        and qualification of \added{robot} behaviors.
        With \pname{Q-Rock}, we suggest a novel, integrative approach to automate robot development
        processes. \pname{Q-Rock} combines several machine learning and
        reasoning techniques to deal with the increasing complexity in the design of robotic
        systems.
        The \pname{Q-Rock} development cycle consists of three complementary processes: 
        (1) automated exploration of capabilities that a given robotic hardware provides, 
        (2) classification and semantic annotation of these capabilities to generate more complex behaviors, 
        and (3) mapping between application requirements and available behaviors.
        These processes are based on 
        a graph-based representation of a robot's structure,
        including hardware and software components.
        \added{A central, scalable knowledge base 
        enables collaboration of} robot designers including mechanical, electrical \added{and systems} engineers, 
        software developers and machine learning experts.
        In this paper we formalize \pname{Q-Rock}'s integrative
        development cycle and highlight its benefits with a proof-of-concept implementation and a use case demonstration.
    \end{abstract}

    \section{Introduction}
    \label{sec:introduction}
Modern robotics has evolved into a collaborative endeavor, where various scientific and engineering disciplines are combined to create impressive synergies. 
Due to this increasingly interdisciplinary nature and the progress in sensor and
actuator technologies, as well as computing hardware and \acrshort{AI} methods, 
the capabilities and possible behaviors of robotic systems improved significantly in recent years. 
Along with the greatly enhanced potential to strengthen established application
fields for robotics and unlock new ones,
these developments pose several challenges for developers and users interacting with robotic systems. 
On the one hand, for hardware and software engineers, these technolo\-gical im\-prove\-ments led to an increasing size of the design
space and, hence, development, integration and programming complexity.
Engineers do not only have to deal with technical peculiarities of a rich
variety of different components when constructing a robot.
They also have to develop advanced control strategies and integrate knowledge
from a range of disciplines in order to unlock the full potential regarding a robot's capabilities.

On the other hand, the field of end users of applications for robotic systems widens,
as more complex and versatile robots open up a wealth of novel applications for which robots were unsuited only years ago.
\changed{E5}{In the last couple of years, the usage of robots has increased
    substantially and robots are used in more and more fields of
application~\cite{stone2016artificial,siciliano2016springer}.}
These users will not be interested in the detailed construction of hardware or software,
but will rather evaluate a robotic system by its possible behaviors and the tasks it can accomplish.
However, without the domain knowledge of an experienced roboticist or AI researcher,
designing a robot and the algorithms that provide the desired behaviors is next to impossible,
and employing engineers for construction of a custom robot is likely to be prohibitively expensive.
\changed{E5}{
Hence, especially small and medium-sized companies face difficulties to adopt robotic 
systems that suit their specific applications~\cite{robot_roadmap}.
To overcome this problem, new robot engineering methods are required.}

We claim that both collaborative teams of roboticists and end users would greatly benefit from a unifying automated 
framework for robot development that spans several abstraction levels to interact on. 
\changed{R1.3,R1.4}{Our initial
conceptual idea is outlined in
\cite{Roehr:2019:IntroducingQRock}, where we introduce \pname{Q-Rock} as
a development framework leveraging integrative
\acrshort{AI}, which means that it applies and combines different \acrshort{AI} technologies in one
system to address the development change. With \pname{Q-Rock} we intend to simplify and automate the whole process 
from robot design to behavior acquisition and its final deployment for experienced roboticists from different disciplines and naive users alike.
In the following sections we detail the implementation of the concept and explain essential elements.
}

\changed{R1.1,E2}{The core hypothesis underlying this project is that the set of all possible
behaviors of a robotic system is \emph{inherently} defined by its constituting hard- and software,
and, furthermore, that this set can be found by self-exploration of the robotic
system.
The major challenge is that the size of the behavior space is subject of the
curse of dimensionality. In order to make the behavior space more manageable, we
restrict the automated exploration to the kinematic capabilities of the robot.
The \pname{Q-Rock} development cycle allows the usage of existing behavior
components from other methods,\added{ e.g.,} for the interpretation of sensor input,
thus permitting a complementary, semi-automated exploration approach. For the
kinematic exploration, we make the assumption that the robot is explored in a
minimal environment that allows to transfer explored capabilities to more
complex environments. Additionally, we limit the capabilities to movements
having a fixed length and we consider only kinematic states that the robot can
reach through the movements itself. These simplifications still cannot negate
the curse of dimensionality. At the lowest level, the robot, due to sensor
resolutions and digital input signals, is a discrete system. Treated as such, a
robot offers a number of possible states and transitions between states, which
is not tractable for higher degrees of freedom in practical applications
\cite{Wiebe2020Combinatorics}. Therefore, in practice we are not able to simulate and store every possible movement individually. Instead, we construct a representative set of movements generated from a parameter space, which encodes all considered capabilities.}

A distinct feature of our approach is that the self-exploration of the hardware
is as goal-agnostic as possible, such that novel behaviors can be synthesized
from already explored capabilities of the system without having to re-perform exploration with a novel task in mind.
This reusability is made possible by clustering capabilities and describing resulting behaviors in a semantically
annotated latent feature space. 

\changed{R1.1,E2}{One important reuse of the capability clusters is the
exploration of systems of systems. Here, the explored capabilities of the
subsystems are used to efficiently explore the capabilities of the assembled new
systems. Even though the exploration of simple subsystems can be already
expensive, the general idea is to considerably increase sample efficiency for
complex assembled systems. This paper deals with the exploration of
base systems, which are required for the capability exploration of hierarchical
systems. The latter, however, will be investigated in our future work.}

Using a growing common knowledge base that links various description levels, from
technical details of single components to behavior classification of
self-explored robotic systems, we also provide a basis for behavior transfer
between systems and reasoning about a possible robotic behavior given its
composition.
Hence, we propose a development cycle with multiple entry points that simplifies and speeds up
the overall design process of robotic systems to benefit both developers and users.

\subsection{Contributions}
The main contribution of \pname{Q-Rock} is the integration of several different subdisciplines of \acrshort{AI} into a common framework 
in order to explore and qualify robotic capabilities and behaviors. 
To this end, we integrate state-of-the-art methods and develop new approaches in four key areas:
\begin{enumerate}[(i)]
	\item Assembly of mechanical, electronic and software components with well defined interfaces and constraints
	\item Exploration and clustering of capabilities aided by machine learning
	\item Ontology-based semantic annotation of behaviors with user feedback
	\item Reasoning about a possible behavior given a robot's hard- and software composition
\end{enumerate}
\changed{R1.2}{In this paper we introduce the formal concepts behind \pname{Q-Rock} and
present a practical evaluation of our approach to tackle robot
design problems in a combination of bottom-up and top-down solving.
The evaluation is based on a \emph{reach} behavior of a robotic manipulator, a mobile
base, and a combination of both.
Developing a simple behavior such as \emph{reach} clearly does not justify a complex
development cycle as presented in this paper.
The implementation of this behavior, however, permits us to (a) outline
a novel, integrative approach to develop robotic systems, which we refer to as the
Q-Rock development cycle and (b) to provide a qualitative analysis of each stage of this
cycle.
Hence, the analysis of the \emph{reach} behavior serves to illustrate the essentials of
the concept, and points to the potential of the \pname{Q-Rock} development
cycle to achieve the automated exploration and classification of robot behavior.
}

\changed{R1.2}{
The \pname{Q-Rock} development cycle supports and simplifies the interaction between
different crafts in multiple ways.
The traditional approach - with often strictly separated and sequenced steps of
either hardware- or software-focused development - should be overcome. 
Firstly, by simplifying the robot design by means to reuse once designed
components - where the link between hardware and software components is already
established.
Secondly, by an automated exploration of newly assembled systems,
whose behaviors cannot always be inferred just from its components.
We outline an approach that accelerates and simplifies robot design processes and
permits robot developers and users to focus on optimization and fine-tuning.}

\changed{R1.2}{
Key elements of the \pname{Q-Rock} development cycle are:
\begin{inparaenum}[(a)]
\item a centralized data and knowledge base,
\item a procedure for automated system analysis by exploring a system assembly's
    possible behaviors (limited by some of our assumption), and
\item semantic annotation of these behaviors to enable their use in reasoning
    procedures.
\end{inparaenum}
}
An important point to note is that \pname{Q-Rock} relies on a well-defined \emph{robot hardware design 
process}, \changed{E3}{which is based on a data and knowledge base} that provides the information to couple hardware and software components 
automatically by specifying requirements and compatibilities. We see this part as a crucial pre-requisite to
implement the \pname{Q-Rock} concept, for which some foundations were developed in the 
precursor project \pname{D-Rock} \cite{DROCK:2018}.
\changed{E3}{Details of the robot hardware design process and its implementation are discussed in Section~\ref{sec:modelling_robot_composition}}

As already stated, \pname{Q-Rock} combines different fields of research, where each area might
have its own interpretation and definition for the same term.
Where needed, in order to avoid confusion through conflicting connotations of a term, we
decided to introduce a new one instead.

\subsection{Paper Outline}
In the following Section~\ref{sec:sota}, we give a short overview of the
current state and limitations of automatic robot behavior learning.
Section~\ref{sec:concept} introduces the concept of \pname{Q-Rock}, provides an overview of the 
methodology and formally defines the procedures and abstractions to implement \pname{Q-Rock}.
Section~\ref{sec:use_case} describes exemplary use case scenarios to
illustrate the implemented concepts presented in this paper.
A discussion in Section~\ref{sec:discussion} concludes the paper.

    \section{State of the Art}
    \label{sec:sota}
    Multiple disciplines, i.e., knowledge \added{representation and }reasoning, task planning as well as machine learning,
can provide important methods to explore robotic capabilities or combine capabilities to generate more complex ones.  
However, to the best of our knowledge, there is little work in automated robot design that is approached in a holistic 
way as it is done in \pname{Q-Rock}. We will highlight relevant holistic approaches here, whereas related work in subdisciplines of the
\pname{Q-Rock} development cycle are described in the corresponding paragraphs.

Ha et al.~\cite{ha2018computational} suggest an automated method for the design
of robotic devices using a database of robot components and a set of motion
primitives.
They use a high level motion specification in the form of \added{end effector} waypoints
in task space.
Their system then takes this motion specification as input and generates the
simplest robot design that can execute this user-specified motion.
However, Ha et al. do not consider the inverse problem which is the base concept of \pname{Q-Rock}:
\changed{E4}{finding all motions a device can perform - under given assumptions.}

A similar development, tackling the problem of learning motion behaviors
via exploration is pursued in the project \pname{memmo} (Memory of Motion) 
\cite{mansard2018using},
where a graph in state space is generated during exploration, and where the links
between nodes refer to control strategies adhering to the
system dynamics. Both graph and control strategies are refined during exploration,
and the resulting trajectories are then used during deployment to warm-start
an optimal control framework.
The key difference to our approach is that in the \pname{memmo} framework,
task objectives need to be known and encoded
in a loss function for training, whereas our framework is mostly goal agnostic during exploration.

A system providing access to robotics development via a web based platform is
included in the \gls{AWS} \cite{AWS:2020}. The services include a RobotMaker
which basically enables use of \gls{ROS} based tools via browser windows. This
way the user doesn't have to install any tools locally. \added{However, }as far as it
could be investigated, even though an account could be created freely, most of
the services are commercial. Additionally, even though the \gls{ROS} community
provides many solutions for different applications, a tool that provides an easy
access to a non-expert user, as aimed at by the \pname{Q-Rock} system, is
lacking and is also not provided by \gls{AWS}.

Another holistic approach for constructing and simulating robots is presented by the Neurorobotics 
Platform \cite{neurorobotics}, under development within the Human Brain Project 
\cite{humanbrainproject}. At the time of writing, this \added{web-based} framework includes an
experiment designer, robot construction for simple toy robots (Tinkerbots \cite{tinkerbots}), a range of predefined 
robots and brain models, and various plotting and visualization tools. The focus lies on 
fostering collaboration between neuroscientists and roboticists and providing simulated 
embodiment for biologically inspired brain models. In \pname{Q-Rock} we rather focus on 
exploration of possible capabilities given a robot's composition, and linking these 
capabilities and corresponding behaviors to its properties.


    \section{\pname{Q-Rock} Development Cycle}
    \label{sec:concept}
\begin{figure*}
    \centering
    \captionsetup{type=figure}
    \includegraphics[width=0.9\linewidth]{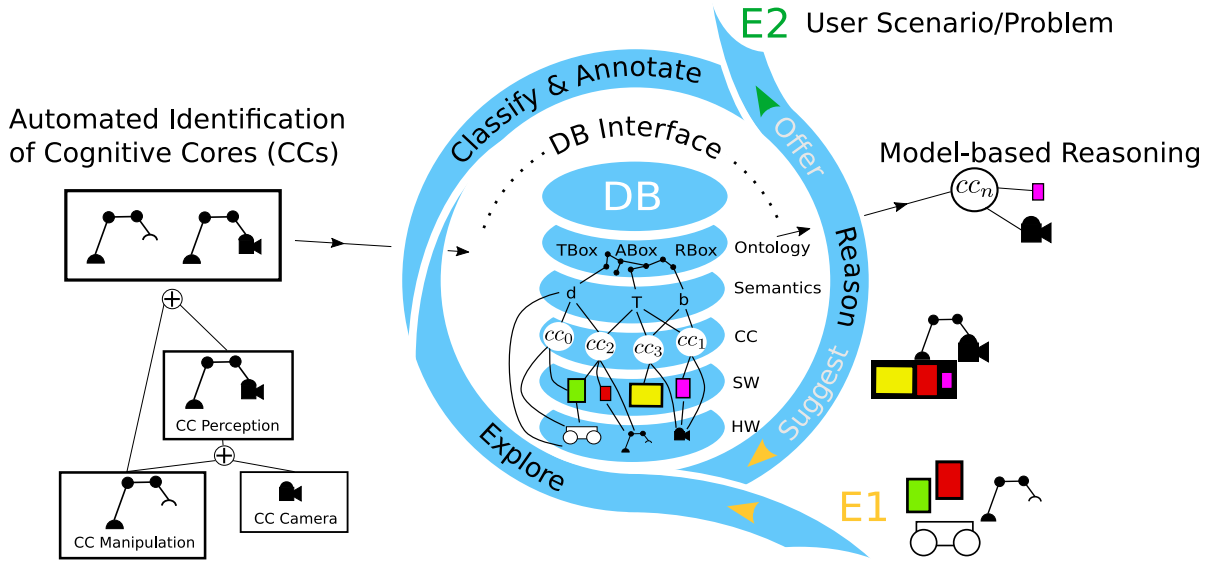}
    \caption{The \pname{Q-Rock} development cycle consists of three complementary steps: 
    ``Exploration'', ``Classification and Annotation'', and ``Reasoning''. 
    A graph database serves as a central knowledge representation and data exchange hub. 
    The process may be initiated from two entry points (E1 and E2), 
    depending on the intention of a user. 
    The entry point specifies whether \pname{Q-Rock} follows a \emph{bottom up} (E1) 
    or \emph{top down} (E2) development approach. 
    %
    }
    \label{fig:qrock_cycle}
\end{figure*}

    To explore and annotate the inherent capabilities and possible behaviors of a 
robot and subsequently allow for reasoning about relations between composition 
and behaviors, \pname{Q-Rock} combines different kinds of \acrshort{AI} techniques in a 
\emph{development cycle} (see Fig. \ref{fig:qrock_cycle}). 
This cycle can be driven by the high-level task specifications 
of a user, but is also flexible enough to support experienced domain experts.
The cycle is divided into three major steps: 
(i) simulation-based exploration of the capabilities of a given piece of robot
hardware,
\changed{R1.14}{(ii) clustering and annotation of these capabilities to generate behaviors and behavior models, 
	and (iii) model-based reasoning about the set of behaviors and related hard- and software that is 
	required for a specific task.}
\changed{E3}{The \pname{Q-Rock} database - implemented as a combination of hand
    curated ontologies~\cite{Korcut:2021} and a graph database\footnote{https://janusgraph.org} - provides the central knowledge base to connect all steps. 
The database provides information about known hard- and software
components and their relations, e.g., compatibility of component interfaces, and the structure of available robotic systems,
As central storage of the results of each stage, the database enables the
immediate use of data across all workflow steps, which leads to a fully integrated development workflow.
}
The development cycle can be initiated from
two entry points (E1 and E2 illustrated in Figure~\ref{fig:qrock_cycle}). 
The first entry point E1 allows to enter a bottom-up development approach. 
Here, the goal is to identify the capabilities of a given robotic system or subsystem.
E1 starts with hard- and software composition and ends up with 
all capabilities of that system, organized in semantically described \acrlongpl{CC}.

The second entry point E2 represents a top-down approach. 
A user triggers the development cycle by providing a task definition, i.e., 
a given user scenario consisting of an environment and a specific problem that a
robot, which is not known to the user, shall solve. 
The goal is to either find a robot in the database that is suitable to address
the specified problem
or to suggest a novel composition that will likely solve the task.

\begin{figure*}
    \centering
    \captionsetup{type=figure}
    \includegraphics[width=\linewidth]{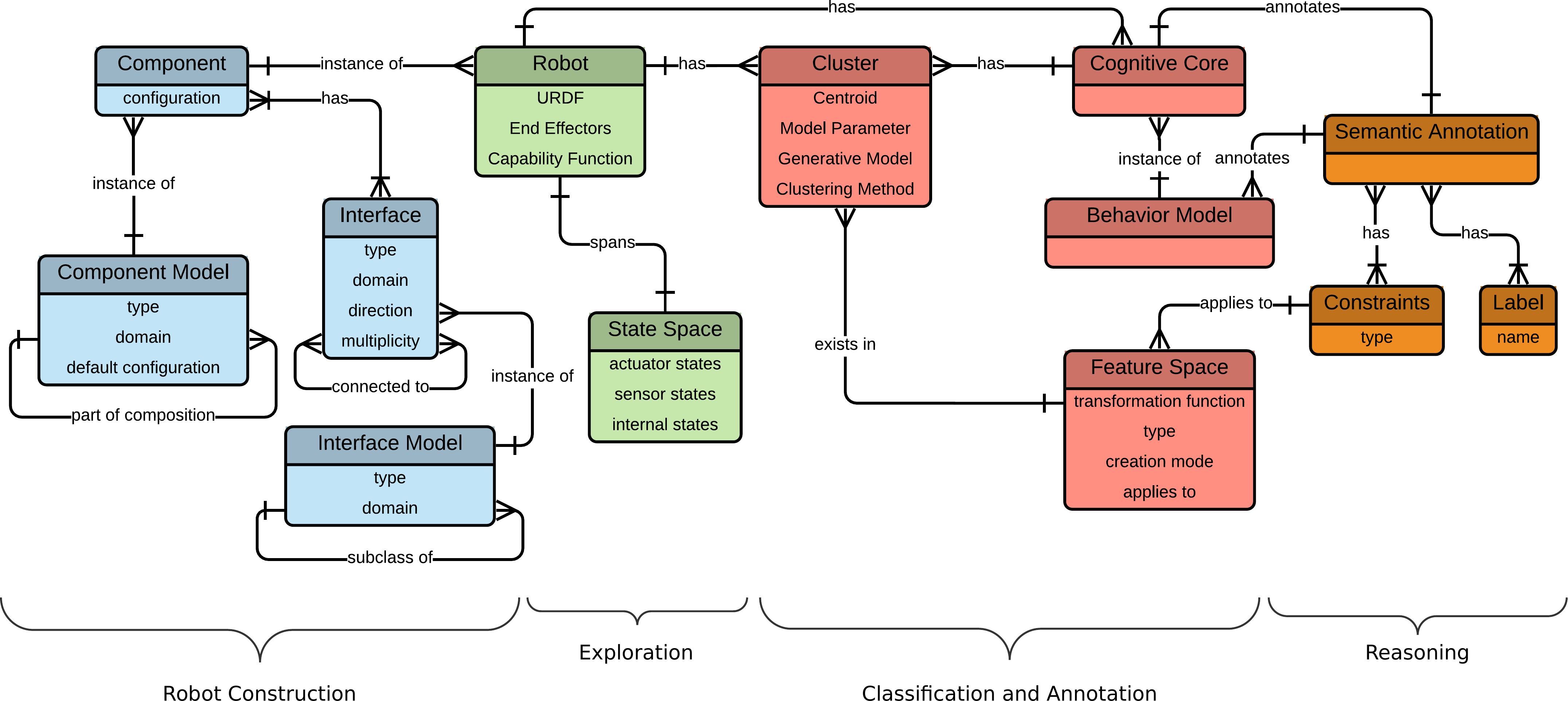}
    \caption{Full \gls{ER} diagram of the \pname{Q-Rock} database. Cardinality symbols
        conform to the \gls{UML} standard. Entities mostly involved in and created during
robot composition / exploration / classification / reasoning are colored blue /
green / red / orange respectively.}
    \label{fig:entity_relationship_model}
\end{figure*}

Complementary to the \pname{Q-Rock} development cycle overview in
Figure~\ref{fig:qrock_cycle}, we provide a standard \acrlong{ER}
diagram in Figure~\ref{fig:entity_relationship_model} to illustrate
involved entities and their relationships.
The following sections motivate and outline the different steps of the \pname{Q-Rock}
development cycle, and successively introduce these entities and their definitions to formalize our approach.

    \subsection{Modelling Robot Composition}
    \label{sec:modelling_robot_composition}
    For all steps of the development cycle 
it is essential to have a well-defined model of a robotic system.
In \pname{Q-Rock}, we represent a robotic system, i.e., the specific types and compositions thereof - as 
well as relations between - robot hard- and software components, using a graph-based
model.

\subsubsection{Related Work}
The formal \gls{AADL} is designed to describe both processing and communication resources, as well as software components and their dependencies.
A system designer is supposed to thoroughly model the system design,
such that given an application designed by the application designer, it can be deployed on the system.
Furthermore, it is possible to use special tools to perform design analysis prior to compilation and/or testing in order to find errors before deployment.
A detailed overview of \gls{AADL} can be found in \cite{AADL2004}.

TASTE is a framework developed by the European Space Agency to design, test and deploy safety-critical applications.
Is uses \gls{AADL} as the modelling layer to design systems and applications.
Based on these models the framework builds the glue code and enables the deployment of the software to a variety of different processing and communication infrastructures.
Details can be found in \cite{Taste2012}.

In contrast to the aforementioned approaches, the domain-specific language NPC4 developed by Scioni et al.~\cite{Scioni2015} uses hypergraphs to model all aspects of structure in system design, software design and other domains.
Its four main concepts are node (N), port (P), connector (C) and container (C) combined with the two relations contains (C) and connect (C); refinements of these concepts form domain-specific sublanguages.
A detailed description of the concept and the language NPC4 is presented in \cite{Scioni2015}.

Our approach aims at exploiting the flexibility and formalization of NPC4 for a structural reasoning approach and combine it with the well-known and tested concepts of TASTE/AADL.
However, unlike NPC4 our approach is based on standard graphs to make use of \added{state-of-the-art} database technology.

\subsubsection{Approach \& Formalization}
\paragraph*{Components}
Components represent the hard- and software building blocks of robotic systems, 
which can be combined to generate more complex components.
Hence, a hierarchy of components of different complexity is created.
At the lowest level of this hierarchy are \emph{atomic components}, 
which can not be divided into other components in our model.

Components are grouped into a predefined, but extendable set of domains
$\mathbb{\mathcal{D}}=\{\mathcal{S},\mathcal{P},M,\mathcal{E},\mathcal{A}\}$.
The domains are described in Table~\ref{tab:domains}.
\begin{table}[h]
\centering
\captionsetup{type=table}
\caption{Predefined domains for components in a robotic system}
\label{tab:domains}
\begin{tabular}{llp{3.5cm}}
\textbf{Name} & \textbf{Symbol} & \textbf{Description} \\\toprule
\textbf{Software}    &   $\mathcal{S}$ & Software components which are, individually or in combination, used to interface and control a robotic system \\
\textbf{Computational} &   $\mathcal{P}$ & Physical entities which represent computational resources and the communication infrastructure \\
\textbf{Mechanics}   &   $\mathcal{M}$ & Components required to establish a mechanical structure / kinematic design of a robotic system \\
\textbf{Electronics} &   $\mathcal{E}$ & Electronic devices including sensors, actuators, power supplies and supply lines \\
\textbf{Assembly}    &   $\mathcal{A}$ & Composite components comprised of any set of components of the previously mentioned domains \\\bottomrule
\end{tabular}
\end{table}
Each domain can only form new components by combining other components of the same domain (unless they are \emph{atomic} components).
The only exception is the \emph{Assembly} domain which allows the composition of components of different domains.
Thus, the \emph{Assembly} domain is the one in which complete robotic systems - 
including their mechanical, electrical, processing and software structure - can be represented.

The main entities and their relationships are represented as labelled vertices 
and edges in a graph $G = \left( V,E,s,t,\Sigma,p_v,p_e \right)$.
Here, $V$ is the vertex set, $E$ the edge set with $s,t$ identifying source and
target vertices, $\Sigma$ is a vocabulary, and $K \subset \Sigma
\backslash \{\emptyset \}$ is a set of predefined keys with ${ 'label' } \in K$.
Property functions for vertices and edges are defined as $p_v: V \to K \times \Sigma$ and
$p_e: E \to K \times \Sigma$, where $p_v(\text{'label'}) \neq \emptyset$ and
$p_e(\text{'label'}) \neq \emptyset$.
The main entity sets are listed in Table~\ref{tab:entities}, whereas 
relations are listed in Table~\ref{tab:relations}.
Note that all entities are represented as vertices in the graph, so that all
entity sets listed in Table~\ref{tab:entities} are subsets of the vertex set
$V$, and likewise all relations are subsets of the edge set $E$.
\begin{table}[h]
\centering
\captionsetup{type=table}
\caption{Entities for modelling a robot's structure}
\label{tab:entities}
\begin{tabular}{llp{3.5cm}}
\textbf{Name} & \textbf{Symbol} & \textbf{Description} \\\toprule
\textbf{component model} & $M_\mathcal{C}$ & All available component models in the respective domains \\
\textbf{component}       & $\mathcal{C}$ & Instances of component models \\
\textbf{interface model} & $M_\mathcal{I}$ & All interface types a component (model) could expose \\
\textbf{interface}       & $\mathcal{I}$ & Instances of interface models - possibly owned by some component (model) \\
\bottomrule
\end{tabular}
\end{table}

\begin{figure}
    \captionsetup{type=figure}
    \includegraphics[width=\linewidth]{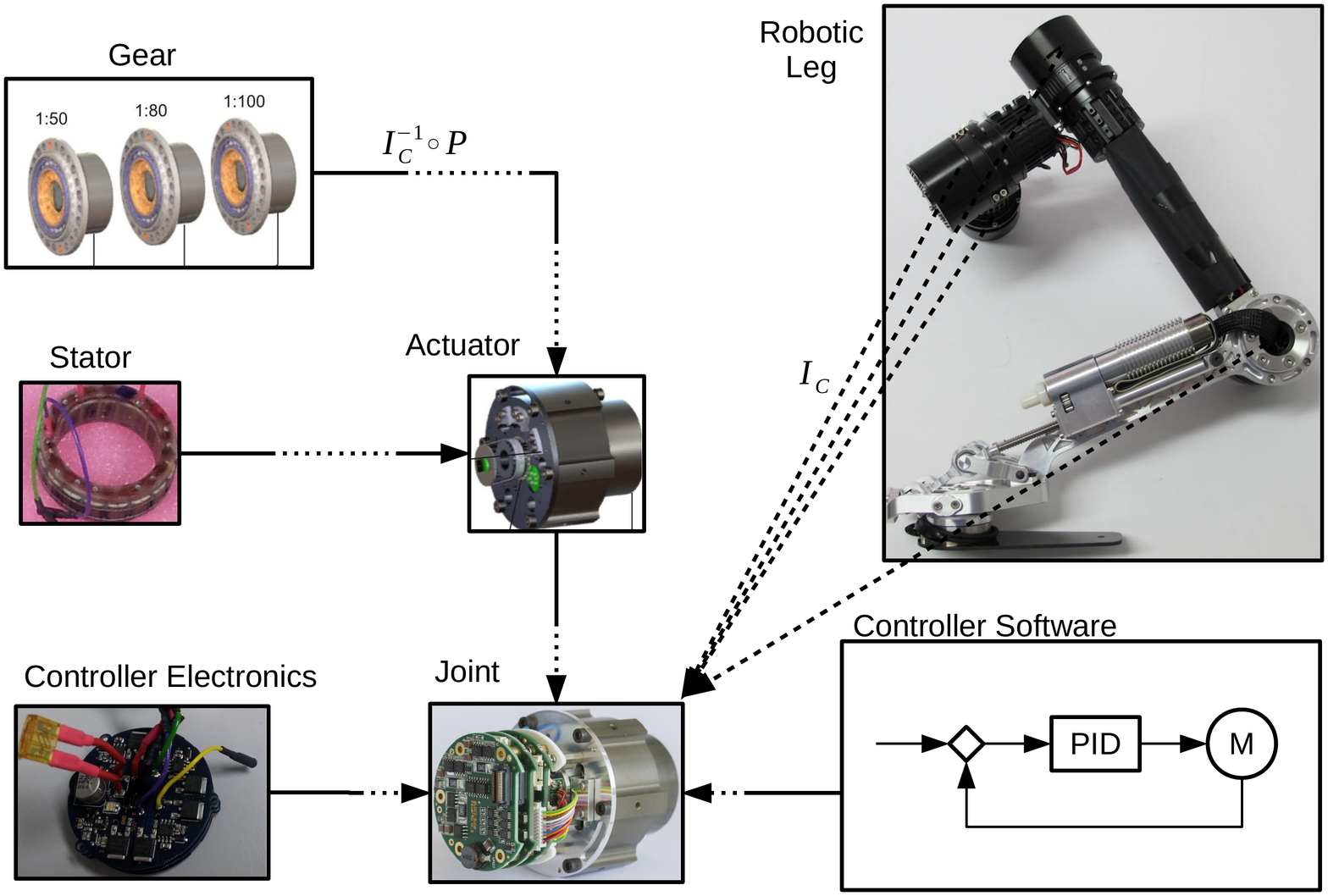}
    \caption{Schematic overview of components of different domains composed to form a higher-level 
    robotic component. Here, gears and rotor/stator components form an actuator. The actuator, the 
    controlling electronics and software components form a joint which can be used to define a robotic leg.}
    \label{fig:component_overview}
\end{figure}

\begin{table*}[h]
\centering
\captionsetup{type=table}
\caption{Available relation types to describe relationships between entities}
\label{tab:relations}
\begin{tabular}{llcp{7cm}}
    \textbf{Relation Name}  & \textbf{Definition} & \textbf{Cardinality} & \textbf{Description} \\\toprule
    {instance of (component model)} & $I_\mathcal{C}: \mathcal{C} \to M_\mathcal{C}$ & N:1 & specify the model $x \in M_\mathcal{C}$ a component $y \in \mathcal{C}$ has to comply with \\
    {instance of (interface model)} & $I_\mathcal{I}: \mathcal{I} \to M_\mathcal{I}$ & N:1 &  specify the model $x \in M_\mathcal{I}$ an interface $y \in \mathcal{I}$ has to comply with \\
    {subclass of}                   & $S: M_\mathcal{C} \to M_\mathcal{C}$ & N:1 &  establish a hierarchy among the component models \\
    {(model) has interface}         & $H:M_\mathcal{C} \to \mathcal{I}$ & 1:N & define the interfaces a component model exposes to the exterior \\
{(component) has interface}     &  $H_\mathcal{C}:\mathcal{C} \to \mathcal{I}$ & 1:N & define the interfaces a component exposes to the \\
                                &                                             & & exterior, given the model of it as a template \\
{part of composition}           &  $P: \mathcal{C} \to M_\mathcal{C}$ & N:1 & partially define the inner composition of components of other components \\
{connected to}                  &  $C_o: \mathcal{I} \to \mathcal{I}$ & M:N &  connect components to form networks, thus refining the inner composition \\
{alias of}                      &  $A: \mathcal{I} \to \mathcal{I}$ & 1:1 &link interfaces in the interior to the exterior of a component model \\
\bottomrule
\end{tabular}
\end{table*}
%
Relations have to be constrained to form a consistent system.
The relations $I_{\mathcal{C}}, I_\mathcal{I},P,S$ are many-to-one relations; 
that means, that no element of their domain can be mapped to more than one element in their co-domain.
This constraint ensures, for instance, that parts of one component model cannot be parts of another component model.
The relations $H, H_\mathcal{C}$ are one-to-many relations, 
thus preventing ambiguity of interfaces between different entities.
$C_o$ is a many-to-many relationship, allowing any connection between interfaces, 
whereas the $A$ relation is a one-to-one relation between an external and an internal interface.
\begin{table*}[h]
\centering
\captionsetup{type=table}
\caption{List of available operators, which allow to modify the graph structure}
\label{tab:operators}
\begin{tabular}{llp{11cm}}
\textbf{Name} & \textbf{Definition} & \textbf{Description} \\\toprule
${create}_\mathcal{C}$    & $\mathcal{D} \times \Sigma \to M_\mathcal{C}$    & add a component model to the graph \\
${create}_\mathcal{I}$    & $\mathcal{D} \times \Sigma \to M_\mathcal{I}$    & add an interface model to the graph \\
$instantiate_\mathcal{C}$ & $M_\mathcal{C} \times \Sigma \to I_\mathcal{C}$  & create a component $c$ from a model $m$ s.t.  $\left(c,m\right) \in \mathcal{I}_\mathcal{C}$\\
$instantiate_\mathcal{I}$ & $M_\mathcal{I} \times \Sigma \to I_\mathcal{I}$  & create an interface $i$ from an interface model $m$ s.t. $\left(i,m\right) \in I_\mathcal{I}$ \\
$isA$                     & $M_\mathcal{C} \times M_\mathcal{C} \to S$ & make component model $x \in M_{\mathcal{C}}$ a subclass of component model y $M_{\mathcal{C}}$, such that $\left(x,y\right) \in S$ \\
$has_M$         & $M_\mathcal{C} \times \mathcal{I} \to H$ &
associate an interface instance with a component model\\
$has$                     & $\mathcal{C} \times \mathcal{I} \to H_\mathcal{C}$ &
associate an interface instance with a component\\
$compose$                 & $\mathcal{C} \times M_\mathcal{C} \to P$ &
make a component part of a component model\\
$connect$                 & $\mathcal{I} \times \mathcal{I} \to C_o$ &  define a
connection between two interface instances \\
$export$                  & $\mathcal{I} \times \mathcal{I} \to A$ & define one
interface to be an alias for another interface, e.g., to map a component model's
interface to a composing component's interface\\
\bottomrule
\end{tabular}
\end{table*}
%
Given the entities and their relations, the operators listed in
Table~\ref{tab:operators} are defined on the graph.
Algorithm~\ref{alg:example1} serves as example to illustrate the usage of these
operators to construct the component model of a robotic leg.
It is assumed, that a component model for a robotic joint $J \in M_\mathcal{C}$
with two (external) mechanical interfaces $a,b \in \mathcal{I}$ exists.
Furthermore, the existence of a component model for a robotic limb $L \in
M_\mathcal{C}$ with two (external) mechanical interfaces $x,y \in \mathcal{I}$
is assumed.
\begin{algorithm}
    \begin{algorithmic}
        \State $j_1 \gets instantiate_\mathcal{C}(J, \mathrm{hip1})$
        \State $j_2 \gets instantiate_\mathcal{C}(J, \mathrm{hip2})$
        \State $j_3 \gets instantiate_\mathcal{C}(J, \mathrm{hip3})$
        \State $j_4 \gets instantiate_\mathcal{C}(J, \mathrm{knee})$
        \State $connect(j_1.b, j_2.a)$
        \State $connect(j_2.b, j_3.a)$
        \State $l_1 \gets instantiate_\mathcal{C}(L, \mathrm{upperLimb})$
        \State $l_2 \gets instantiate_\mathcal{C}(L, \mathrm{lowerLimb})$
        \State $connect(j_3.b, l_1.x)$
        \State $connect(l_1.y, j_4.a)$
        \State $connect(j_4.b, l_2.x)$
        \State $m = create_\mathcal{C}(\mathcal{A}, \mathrm{Leg})$
        \State $compose(j_1, m)$
        \State $compose(j_2, m)$
    \end{algorithmic}
    \caption{Example application of graph operators to constructing a component model}
    \label{alg:example1}
\end{algorithm}
%
Figure~\ref{fig:component_overview} visualizes the graph structure resulting
from running Algorithm~\ref{alg:example1}.
The graph has three components $g_i$, gears of different ratio, in the mechanical domain $g_i \in M$.
One of it has been instantiated ($I_\mathcal{C}$) and is part of ($P$) an
actuator $A \in \mathcal{A}\cap M_\mathcal{C}$. Chaining the respective
relations $I_\mathcal{C}^{-1} \circ P$ (see Table~\ref{tab:relations}) resolves
to:
\[
\left\{ (g_i,a) | \exists x \in \mathcal{C} : (g_i,x) \in I_\mathcal{C}^{-1} \land (x,a) \in P \right\} \text{for some }i.
\]
The component (instance) actuator $a$ with stator and gear as its composing parts
is combined with controller electronics and controller software to define the
joint model.
This model defines the structure of joint instances in the higher-level leg component.

\changed{E3}{
\subsubsection{Implementation}
The robot hardware design process presented in this section has been implemented
using a state-of-the-art graph database.
The database directly supports the creation of a graph $G$ with labelled vertices and edges and - most notably - the direct specification of cardinality constraints depicted in Table~\ref{tab:relations}.
The database is accessed at low-level via the \emph{Gremlin query
language}~\cite{2021:Gremlin} and a higher-level through a Python layer that we
implemented.
The higher-level Python layer translates (sub-)graphs of the database into a
network of interconnected Python objects.
It is also this layer in which the graph operators of Table~\ref{tab:operators}
are implemented as the basis for the whole \pname{Q-Rock} toolchain.
}

    \subsection{Exploration}
    \label{sec:exploration}
    The exploration step in the \pname{Q-Rock} development cycle aims at finding all
capabilities a given piece of robotic hard- and software can provide in a meaningfully
defined exploration environment. The chosen exploration environment has to allow
a transfer to more complex application environments.
Capabilities in this context mean the possible trajectories a robot
can produce within a robot state space and a world state space.
The exploration is based on simulating a robot that has been modelled as
formally described in Section \ref{sec:modelling_robot_composition} - a
practical example is given in Section~\ref{sec:use_case}.

\subsubsection{Related Work}
In distinction to other approaches in robotics,
we try to avoid directing the exploration towards any kind of goal, and
instead aim at generating a maximal variety of capabilities to find a
representative set that might include novel, unanticipated ones.

Capability exploration methods usually aim to create a library of diverse
capabilities~\cite{gregor2016variational,eysenbach2018diversity,achiam2018variational,pathak2019self},
so that the coverage of the behavior space is maximized and the capabilities can be utilized in
different tasks and environments. \added{For} the capabilities to be transferable between tasks,
these approaches avoid task specific reward functions.
Instead they use intrinsic motivations such as novelty, prediction error and empowerment.
An extensive overview of intrinsic motivations in reinforcement learning can be found in \cite{aubret2019intrinsic}.

Because of their inherent incentive to explore and find niches,
evolutionary algorithms are natural candidates for behavior exploration.
Lehman and Stanley~\cite{lehman2011abandoning} propose to use \emph{novelty}
as the sole objective of likewise-called novelty search. It was found to perform significantly
better than goal-oriented objectives in deceptive maze worlds.
Novelty search has already been applied to robotics to find multiple \emph{diverse} high
quality solutions for a single task \cite{cully2018quality,kim2019exploration}.
Cully~\cite{cully2019autonomous} suggests combinations of different methods,
e.g., quality-diversity optimization methods and unsupervised methods, which allow to explore various capabilities of a system without any prior knowledge about their morphology and environment.

\subsubsection{Formalization}
\begin{figure}
    \captionsetup{type=figure}
    \includegraphics[width=\columnwidth]{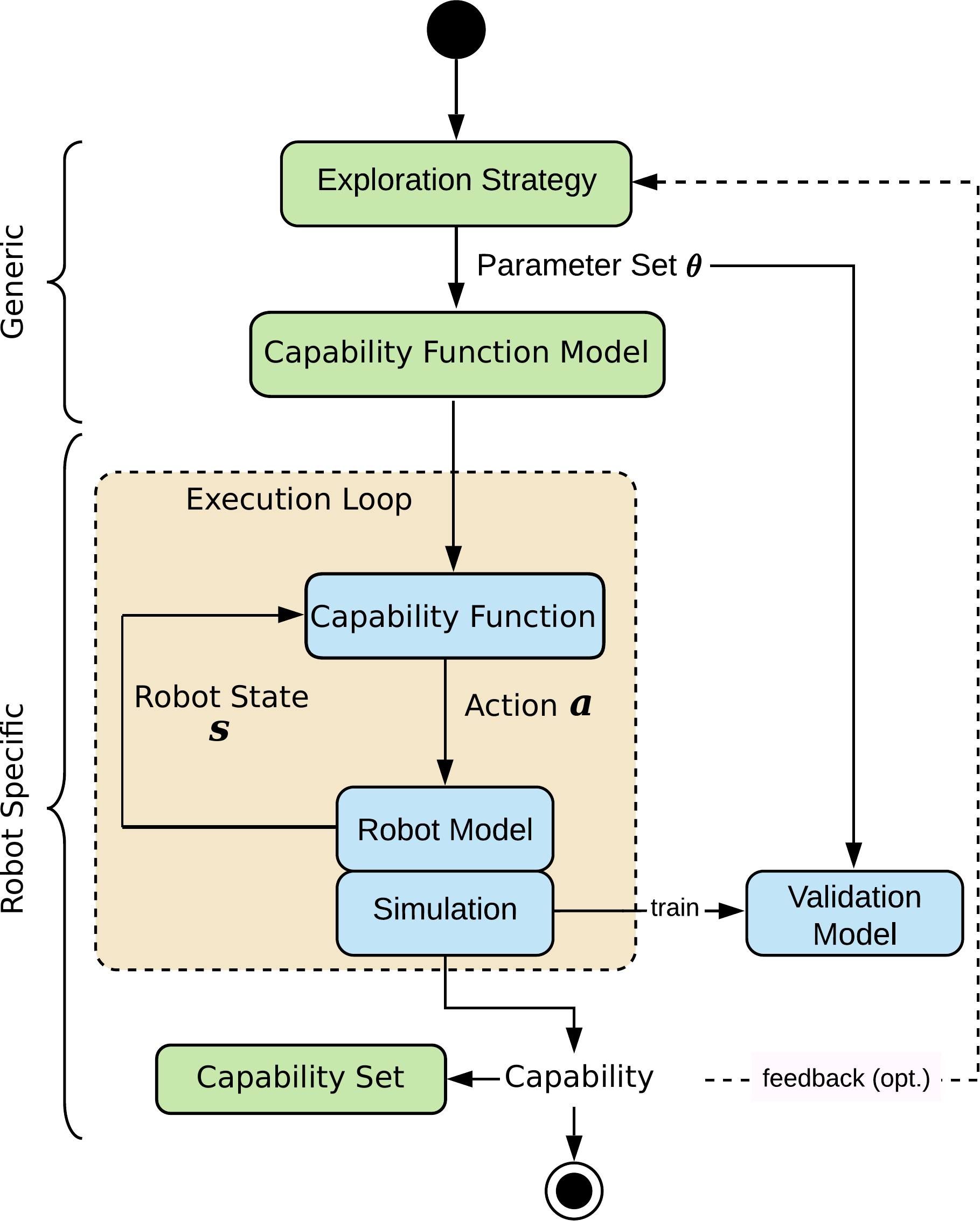}
    \caption{Key elements of the exploration: an exploration strategy generates a
        parameter set which is combined with a \acrlong{CFM} to generate
        a \acrlong{CF}. An execution loop uses the \acrlong{CF} to
generate the actual capability from a simulation, while a validation model is trained in parallel.}
    \label{fig:exploration_overview}
\end{figure}

The abstracted approach of the exploration is depicted in
Figure~\ref{fig:exploration_overview}. It serves as high level description of
the process where the formalization will be given in the coming
paragraphs.
Exploration discovers a set of capabilities by applying a search
strategy, where the challenge lies in handling a significantly large state space.
We tackle the large state space using a parameter-only based encoding for
\acrlongpl{CF}: the encoding is compact and yet arbitrarily precise.
Creating a \acrlong{CF} from a dedicated \acrlong{CFM} and applying it on the actual
robot in an execution loop results in a capability of the system, where a
capability is the executed trajectory in the world state space.
This structure allows to validate the feasibility and cluster
\added{robot-specific} execution characteristics from capabilities on the basis of the input parameter space.

\paragraph*{States}
\begin{definition*}[Joint State]
    The \emph{joint state} $\pmb{q} \in \pmb{Q}$ is a vector of all joint positions of the robot.
    For a robot with n joints: $\pmb{Q} \subset Q^1 \times Q^2 \times \dots \times Q^n$. $\pmb{Q}$ 
    is a subset of the Cartesian product because not all combinations of joint positions may be allowed 
    due to the robot structure.
\end{definition*}

\begin{definition*}[Actuator State]
    The \emph{actuator state} $\pmb{s}^a$ of the robot is a tuple of the
    configuration and the joint velocity, so that $\pmb{s}^a =
    (\pmb{q},\dot{\pmb{q}}) \in \pmb{S}^a$,
    where the actuator state space $\pmb{S}^a = \pmb{Q} \times \pmb{V_{Q}}$
    combines joint state space $\pmb{Q}$ with joint velocity space $\pmb{V_{Q}}$.
    The robot actuator state completely describes the positions and velocities of
    all parts of the robot at a given time.
\end{definition*}
    
The complete (observable) state of the robot contains not only the actuator
state $\pmb{s}^a$ but also the states $\pmb{s}^{s} =
(s^{s,1},\cdots,s^{s,m})\in \pmb{S}^s$ of all $m$ sensors and possibly internal states $\pmb{s}^i$.
\begin{definition*}[Robot State]
    The (full) robot state $\pmb{s}^{rob}$ is a combination of actuator, sensor
    and internal states. An internal robotic state $\pmb{s}^{i} =
    (s^{i,1},\cdots,s^{i,k})$ for k internal properties and $\pmb{s}^{i} \in \pmb{S}^i$ may encompass for example internal time, battery status or a map of the robot's surroundings. $\pmb{S}^a, \pmb{S}^s$ 
    and $\pmb{S}^i$ are the sets of all
    possible actuator, sensor and internal states, respectively. The full robot state reads:
    \begin{equation*}
    \pmb{s}^{rob} = (\pmb{s}^{a},\pmb{s}^{s},\pmb{s}^{i})\in \pmb{S}^{rob} = \pmb{S}^a \times \pmb{S}^s\times \pmb{S}^i.
    \end{equation*}
    The robot state $\pmb{s}^{rob}$ does not contain the complete information about the actual physical or 
    internal state of the robot. It does only contain information that is accessible by the robot itself, 
    i.e., that can be captured. Information about sensorless unactuated joints
    for example is not part of the robot state.
\end{definition*}

\paragraph*{Actions}
To trigger changes of the robot state and thereby generate trajectories of robot states, 
it is necessary to define motor actions. A motor, which is part of a joint, 
outputs a motor torque $\tau \in \mathcal{T}$.
The torques for all joints can be written as a tuple $\pmb{\tau} \in
\pmb{\mathcal{T}} = \mathcal{T}^1 \times \mathcal{T}^2 \dots \mathcal{T}^n$. An
idle joint always outputs $\tau = 0$. 
\begin{definition*}[Action]
    A kinematic \emph{action} $\pmb{a}^{kin}$ is a tuple of a torque $\pmb{\tau}
    \in \pmb{\mathcal{T}}$ and a time interval $\Delta t$, such that
    $\pmb{a}^{kin} = (\pmb{\tau}, \Delta t) \in \pmb{A}^{kin}$, where $\pmb{A}^{kin}$ denotes the kinematic action space.
    Applying a kinematic action to the robot maps the current robot state to a new robot state:
    \begin{equation*}
        \pmb{a}^{kin} : \pmb{S}^{rob} \rightarrow \pmb{S}^{rob}
    \end{equation*}
    Besides kinematic actions, there are also perceptive actions $\pmb{a}^{per} \in \pmb{A}^{per}$ which evaluate sensor data and store abstractions in the internal robot state:
    \begin{equation*}
        \pmb{a}^{per} : \pmb{S}^s \rightarrow \pmb{S}^i
    \end{equation*}
    Finally, there are internal actions $\pmb{a}^{int} \in \pmb{A}^{int}$ processing internal information:
    \begin{equation*}
        \pmb{a}^{int} : \pmb{S}^{i} \rightarrow \pmb{S}^{i}
    \end{equation*}
    The full action space is the Cartesian product of the individual action spaces:
    \begin{equation*}
        \pmb{A} = \pmb{A}^{kin} \times \pmb{A}^{per} \times \pmb{A}^{int}
    \end{equation*}
\end{definition*}

\paragraph*{Environments \& World State}
Note that environments with different properties, including, 
but not limited to, gravitational force, pressure, and
temperature will have an influence on the outcome of a kinematic action.
Hence, environmental parameters as well as poses and properties of objects in
the robot's workspace have to be considered when evaluating the feasibility of kinematic actions.

Furthermore, the environment is an important component to identify certain properties of capabilities. For example a throwing capability relies on the 
temporal evolution of states of the object to be thrown, which is represented by
environment states that can be external to the robot (if it does not have the appropriate sensing capabilities).
This arises also for capabilities that can, at first glance, be considered mostly environment independent: 
The effect of actions on the trajectory of an end effector when pointing is still determined by gravity and the viscosity of the medium in 
which the movement is performed. Even more, there is no generic way to determine the poses of all the robot's limbs just from sensing 
the actuator states $\pmb{s}^a$: if a system is underactuated or does not have
sensors on some actuators, an analytical solution for the feed forward kinematics may not exist.
To compensate for this, we introduce the world state space which may also contain information unavailable to the robot itself:

\begin{definition*}[World State Space]
	The world state $\pmb{s}^{world} \in \pmb{S}^{world} = \pmb{S}^{rob}
        \times \pmb{S}^{obs}$, where the observational state space
        $\pmb{S}^{obs}$ contains states read from the environment, e.g., the
        position and orientation of objects or robot limbs and end effectors.
        These states are obtained during simulation or by monitoring a real
        world execution and will be accessible to the robot if it has the appropriate sensing capabilities.
\end{definition*}


\paragraph*{Capability}
A particular capability will require the sequential execution of a sequence of
actions. Such an action sequence can be represented by a \acrlong{CF}. The
\acrlong{CF} selects an action for the robot based on the current state and
time, and thus defines how the robot is supposed to (re-)act in a given situation. 
\begin{definition*}[Capability Function]
    A \emph{capability function} is a function \emph{cap} that maps the robot state at a given time to an action:
    \begin{equation*}
        \textit{cap}: \pmb{S}^{rob} \times \{0,\dots, T\} \rightarrow \pmb{A}
    \end{equation*}
    and $\textit{cap} \in \pmb{CF}$, where $\pmb{CF}$ denotes the capability function space.
    An important detail to note is that the \acrlong{CF} operates on the robot state
    space and not the world state space. A \acrlong{CF} is a robot inherent function that considers only information that is available to the robot itself.
\end{definition*}

A \acrlong{CF} can be created in various ways, e.g., it could be a policy
obtained from reinforcement learning, a behavior from an evolutionary algorithm, or a control law from optimal control theory. 

In general the generation of \acrlongpl{CF} can be formulated with a \acrlong{CFM}:
\begin{definition*}[Capability Function Model]
    A \emph{capability function model} is a mapping \emph{cfm} from a parameter space $\pmb{\Theta}$ to the capability function space
    \begin{equation*}
        \textit{cfm}: \pmb{\Theta} \rightarrow \pmb{CF}
    \end{equation*}
\end{definition*}

The \acrlong{CFM} introduces a parameter space $\pmb{\Theta}$, which
allows the parametric generation of \acrlongpl{CF} and is the basis for
the exploration. In order to not constrain the exploration, the \acrlong{CFM} should
be able to represent all kinds of \acrlongpl{CF} of a system. In principle, however,
it is also possible to operate with multiple \acrlongpl{CFM} at once.

By repeatedly calling a \acrlong{CF} and applying the resulting actions to the
robot, a capability is executed.
\begin{definition*}[Capability]
    A \emph{capability} $\pmb{l}^T \in \pmb{L}$, where T is the finite time horizon and $\pmb{L}$ the space of all trajectories, is defined by a sequence of world states and time coordinates:
    \begin{equation*}
        \pmb{l}^T=\left[(\pmb{s}^0, t^0), (\pmb{s}^1, t^1), \dots, (\pmb{s}^{T}, t^{T})\right]
    \end{equation*}
    where the transition between successive states $\pmb{s}^t$ and $\pmb{s}^{t+1}$ is effected by an action $\pmb{a}^t$ of the robot.
\end{definition*}

\changed{R1.5}{Capabilities are central entities in the \pname{Q-Rock} philosophy, since 
we argue that a complete set of all possible capabilities is the most
fundamental representation of what a system is able to do. In general the result
of executing a capability strongly depends not only on the robot itself but also on the environment. With the intention to isolate the capabilities of the robot itself, we assume the environment to be minimalistic, deterministic, and static.}


To refine the notion of completeness, we make two crucial assumptions at 
this point:
\begin{assumption}[Discretized Time]
    We assume a \textbf{discretized time} model by arguing that
    (most) robots are controlled by digital hardware or controllers which
    have a specific clock or controller frequency. 
    The smallest time step considered here is the denoted by $\delta t$.
\end{assumption}
\begin{assumption}[Discretized State Space]
    We assume that $\pmb{S}$ is sensed by digital sensors and we consider only state changes if we can distinguish them. As consequence we have a discretized State Space $\pmb{S}$.
\end{assumption}


While this discretization reduces the cardinality of $\pmb{L}$, it is still countably infinite if $t$ is 
not bounded, so we have to choose a maximal capability length T. Now, in principle, a complete set of 
all possible capabilities up to a maximal length can be generated. Not surprisingly, this set would still 
have an intractable size considering typical resolutions of modern hardware and degrees of 
freedom \cite{Wiebe2020Combinatorics}.
A possible approach is to use a generic capability representation instead of
manually specifying a sequence of actions. The parameters of such a
representation define the resulting capability. Possible tools are motion
primitives such as polynomials \added{used by Kumar et al.}~\cite{kumar2016model},
\acrshortpl{DMP} \added{ used by Schaal}~\cite{schaal2006dynamic} or Gaussian kernel functions\added{ used by Langosz}~\cite{langosz2018}.

As a full set of capabilities is not tractable, the next best thing is a representative set of capabilities with a uniform distribution in a given feature space.
An evolutionary algorithm such as novelty search
\cite{lehman2011abandoning} offers a suitable approach. With novelty search it
is possible to search for novel capabilities with respect to a previously
specified characteristic. \added{A possible problem for goal-agnostic exploration however is precisely this characteristic that is 
used to define novelty, since it can bias the distribution of capabilities in an undesirable 
way for subsequent clustering, and partially preempts the distinct feature generation step. 
An alternative strategy is simple random sampling on the parameter space $\Theta$, which also 
comes with its own caveat: The final distribution of the capabilities in later defined feature 
spaces will depend on the specific parametrization. Despite this, we see both approaches as viable alternatives
, since they pose relatively low restrictions on the feature generation compared to more 
goal directed exploration. } A representative set of capabilities, obtained with an
exploration strategy like this, may serve as a starting point for exploring the
space in a finer resolution, for capturing the system dynamics in a model, or for searching for a specific capability.



Because the \acrlong{CFM} itself is \added{robot-agnostic}, it is a priori
not clear, which parameters $\pmb{\theta}$ correspond to feasible capabilities
of the robot. For this reason, a validation model is trained that predicts which
parameters $\pmb{\theta}$
lead to capabilities that are actuable on the robot in the current environment.

After the exploration phase is finished, the obtained library of capabilities,
the \acrlong{CFM} with the simulator to generate new capabilities, and the validation model are saved to the database and can be used by the following step.

    \subsection{Classification and Annotation}
    \label{sec:classify_annotate}
    The goal of this workflow step is the creation of \acrlongpl{CC}. \Acrlongpl{CC} are hubs that connect a specific \acrlong{BM} with 
the robot's hard- and software, \acrlongpl{SA} of that \acrlong{BM}, and
\added{robot-specific} capabilities that execute the behavior. 
\changed{R1.14}{Cognitive cores} allow the execution of the corresponding behavior by using constraints and target values in semantically annotated feature spaces, and rely on clustering of capabilities in these spaces.
Cognitive cores are central entities in \pname{Q-Rock} since they constitute our
solution to the symbol grounding problem, i.e., link semantic 
descriptions to sub-symbolic representations, and serve as a
basis for reasoning about the relation of hard- and software components, robotic structure, and resulting behavior.

\subsubsection{Related Work}
One important point of our approach is the clustering of capabilities into feature 
spaces and the control of the robot within these feature spaces. Several 
studies have shown performance and robustness benefits from controlling a simulated agent in a latent, compressed
feature space. Ha et.al. \cite{ha2018world} used a variational autoencoder on visual input to control a car in a 2D game world. In the context of hierarchical reinforcement learning, Haarnoja et. al. 
\cite{haarnoja2018latent} showed that control policies on latent features outperform 
\added{state-of-the-art} reinforcement learning algorithms on a range of benchmarks, and
Florensa et. al. \cite{florensa2017stochastic} found high reusability of simple
policies spanning a latent space for complex tasks. In a similar vein, Lynch et
al. \cite{lynch2020learning} investigate using a database of play motions, 
\added{i.e.,}
teleoperation data of humans interacting in a simulated environment from intrinsic motivation, combined with projections into a latent planning space to generate versatile control strategies. While the latter study has parallels to our segmentation into an exploration and a clustering phase, no previous approach aims at a semantically accessible feature representation, as we propose in \pname{Q-Rock}. 

A method for generation of a disentangled latent feature space from observations was developed by Higgins et al. \cite{higgins2016early}. This variational autoencoder builds on the classical autoencoder architecture \cite{plaut1987learning}, \cite{hinton2006reducing} that compresses data into a latent space. 
The authors note that disentangling seems to produce features that are also meaningful in a semantic sense, such that changes in feature space lead to interpretable changes in state space. 

\changed{R1.7}{Close to our work, Chen et al. \cite{chen2016dynamic} use a
    combination of variational autoencoders and \glspl{DMP} to 
learn and generate robotic motion. They showcase that semantically 
meaningful representations of motion can be obtained, and that switching between different motions can be performed smoothly. 
To a similar end, Wang et al. \cite{wang2017robust} combine a variational autoencoder with generative adversarial imitation learning and show that a semantic embedding space can be learned for reaching and locomotion behavior.  Whereas both these approaches 
highlight the construction of a semantically interpretable 
latent space 
for motion behaviors, they rely on human training data with a known semantical context. In our framework we aim at extracting the semantics after a motion library has been generated during the exploration step, which has no notion of semantics per se.}

A combination of unsupervised clustering and variational autoencoders is described by Dilokthanakul et al. \cite{dilokthanakul2016deep}. However, direct semantic annotation of these features and a formalized combination into behavior models has not been considered to date.

\subsubsection{Formalization}
To arrive at a formal definition of \acrlongpl{CC}, we first need to clarify what constitutes a behavior. Since the term "behavior" has overloaded definitions in 
various disciplines, we specifically mean behavior in a broad, radical behaviorist sense, while emphasizing the phenomenological aspects: Everything an agent does is a behavior, and all behaviors 
must be in principle completely observable \cite{chiesa1994radical}. The
complete observation is provided by the \acrlongpl{CAP} as defined in Section \ref{sec:exploration}.
We further define a \acrlong{BM} as an abstraction of similar \acrlongpl{CAP} that have the same semantic meaning: A behavior of "walking" is not bound to the exact execution of a sequence 
of robot and world states, but rather a large number of \acrlongpl{CAP} that can differ in certain aspects. 
We thus propose that different \acrlongpl{BM} can be identified by finding
constraints to \acrlongpl{CAP} in appropriate \acrlongpl{FS}, leading to the
following definition:
\begin{definition*}[Behavior Model]
    A \added{robot-agnostic}, semantically labelled abstraction of a set of \acrlongpl{CAP} $L$ that adhere to constraints in \acrlongpl{FS}.
\end{definition*}
\Acrlongpl{FS} arise from transformations of the \acrlongpl{CAP} via a feature function to capture specific aspects, and allow to define distances between \acrlongpl{CAP} within these aspects:
\begin{definition*}[Feature Function]
        A feature function $\text{ff}_k$ maps from capabilities $l \in L$ to a
        set of values in $\mathcal{R}^n_k$, so that $\text{ff}_k: \pmb{L} \to
        \mathcal{R}^n_k$.
	This function is supplemented with a semantic description.
\end{definition*}
\begin{definition*}[Feature Space]
    A metric space $\pmb{F}_k$ with elements $\pmb{f}_k = \text{ff}_k(l)$. It is
    uniquely defined by the combination of $\text{ff}_k$ and its metric $m_k$.
\end{definition*}
An important aspect of the \acrlongpl{FF} is their semantic descriptions, which constitute the language in which the 
\acrlongpl{BM} are defined. 

The \acrlongpl{FF} $\text{ff}_k$ can be obtained in two different ways. Either they are defined manually and directly 
annotated by a semantic description, e.g., \changed{R1.16}{
\begin{equation*}
    \text{ff}_k(l^T) = \frac{1}{T+1}\sum_{t=0}^{T}\dot{\pmb{q}}^t
\end{equation*}
}
with the description 'average actuator velocity'. Alternatively,  they can be found automatically in a purely data-driven way, \added{e.g.,} by using variational autoencoders  \cite{higgins2016early} adapted to trajectory data. The \pname{Q-Rock} framework allows both approaches that can also be used in parallel to 
provide maximal flexibility, whereas the latter is not implemented to date. We thus enable the use of expert knowledge to define the most relevant 
\acrlongpl{FS} for a given problem. However, a non-expert user could also solely resort to the automatic approach. In addition, interesting \acrlongpl{FF} that 
reflect the specifications of the robot model might be discovered automatically
that are not obvious \added{-- even to an experienced observer}, or hard to formulate.

An example of a \acrlong{BM} defined by constraints in \acrlongpl{FS} is:
\begin{align*}
    \mathrm{label}: & \quad \mathrm{reach} \\
    \mathrm{constraints}: & \quad \pmb{F}_1: \mathrm{min:} \, 0.95 \quad \mathrm{max:} \, 1.0 \\
    & \quad \pmb{F}_2: \mathrm{variable} \\
    & \quad \pmb{F}_3: \mathrm{variable}
\end{align*}
Currently, only min/max and variable constraints are implemented. A variable constraint means that a target value $\pmb{f}_k^{tar}$ has to be provided when the 
associated behavior should be executed by a robot. Since \acrlongpl{BM} are
\added{robot-agnostic}, they can be
grounded for different robotic systems. The \added{robot-agnostic} nature of the \acrlong{BM} depends on \acrlongpl{FF}' semantic descriptions: \acrlongpl{FF} having the same effect on a semantic level may have varying definitions for different robots, especially if they are represented by encoder networks or other function approximators. Thus it must be possible to identify \acrlongpl{FF} across robots by their semantic description. 

To achieve a \added{robot-specific} grounding of the \acrlong{BM}, the \acrlongpl{FS} 
$\pmb{F}_0, \dots, \pmb{F}_k$ are populated by mapping \added{robot-specific} \acrlongpl{CAP} 
provided by the exploration step via the associated \acrlongpl{FF}
$\text{ff}_1,\dots,\text{ff}_k$. 

In principle, if all \acrlongpl{CAP} of a robot are contained in the representative set provided by the exploration step, a simple lookup of \acrlongpl{CAP} that adhere to the \acrlong{BM} constraints is sufficient to execute the desired behavior. However, as noted before, this usually implies a capability set of intractable size. 

\begin{figure}
    \captionsetup{type=figure}
    \includegraphics[width=\columnwidth]{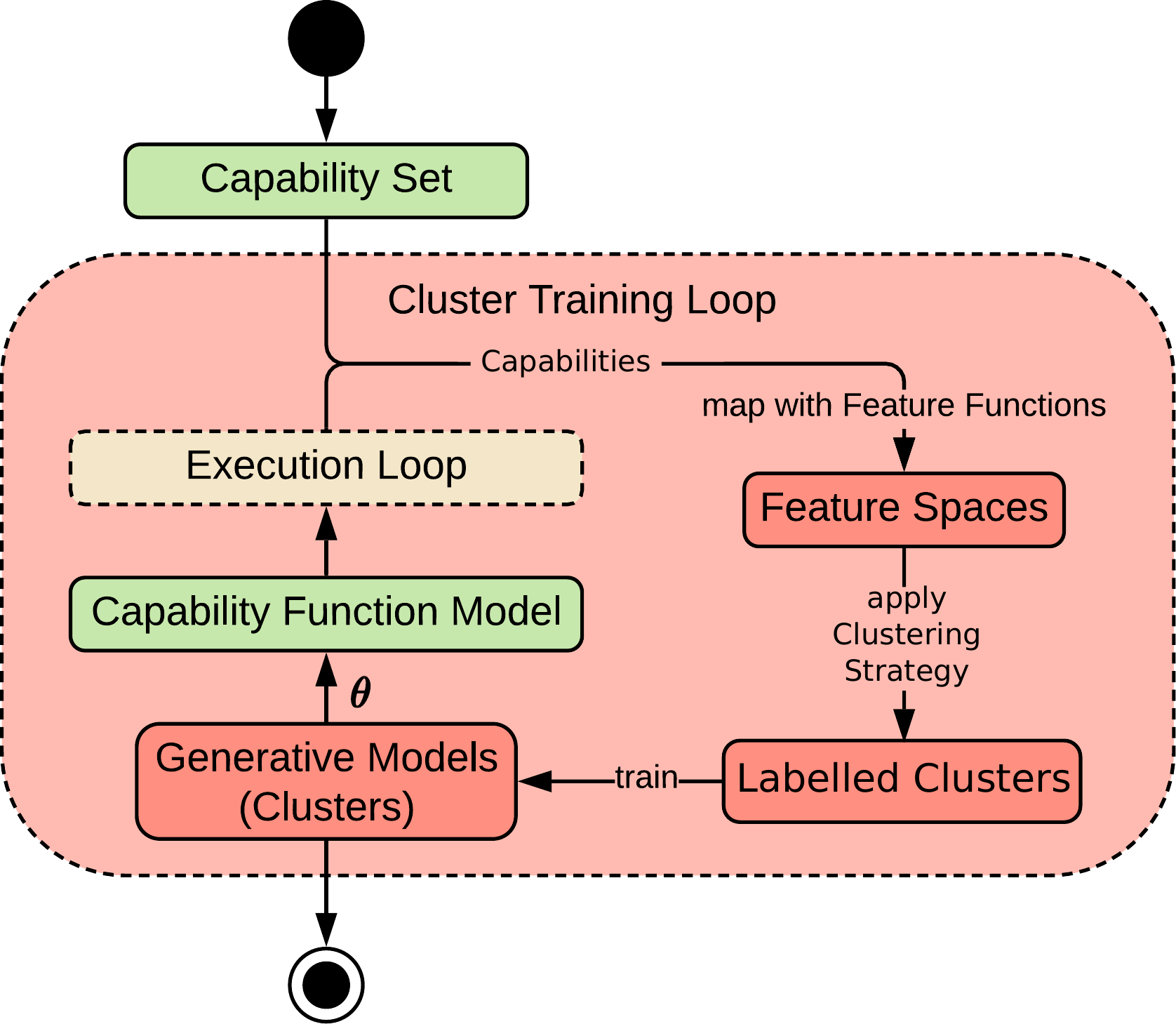}
    \caption{Clustering overview. A representative capability set, along with
        the corresponding parameters $\pmb{\theta}$ and \acrlong{CFM} is
        provided by the exploration. Transformation functions $\text{ff}_k$ are applied to map to 
    \acrlongpl{FS} $\pmb{F}_k$. In these \acrlongpl{FS}, clustering is
    performed. The labelled clusters are used to train probabilistic generative models on
    the parameter 
    space $\pmb{\Theta}$, s.t. clusters can be stored in an efficient and expressive way. When sampling from the generative cluster models, parameters $\pmb{\theta}$ are generated that 
    lead to \acrlongpl{CAP} in the intended cluster. The mapping from parameters
    to a capability is mediated by the \acrlong{CFM} and the execution loop (see Figure
    \ref{fig:exploration_overview}). During training, sampling of parameters and generation of new \acrlongpl{CAP} is used to verify model performance.}
    \label{fig:clustering_overview}
\end{figure}

We tackle this problem in two ways: Firstly, the \acrlongpl{CAP} are clustered
in $\pmb{F}_k$ $\forall k$ and the centroids of the clusters are used to check
constraints for all members of the cluster. Whereas the result of this check is
not exact for all \acrlongpl{CAP}, computational performance is greatly
increased. Secondly, to avoid a lookup search when executing a
behavior and to not be restricted to \acrlongpl{CAP} seen during exploration, we
abstract generative models on the parameter sets $\pmb{\theta}$ from the
capability clusters. Thus, clusters are represented by probabilistic generative
models that, when sampled from, provide parameters $\pmb{\theta}$ which, via recurrent execution of the \acrlong{CFM}, lead to \acrlongpl{CAP} that likely lie in the intended clusters. Clusters are thus defined as:
\begin{definition*}[Cluster]
    A cluster with label $c_k^j$ is defined within a \acrlong{FS} $\pmb{F}_k$,
    which is associated with several clusters $j \in [1, n_k]$, where $n_k$
    denotes the number of clusters found in $F_k$.
    Each cluster has a generative model $G_k^j(\pmb{\theta}) \approx
    p(\pmb{\theta}, c_k^j) = p(\pmb{\theta} | c_k^j)p(c_k^j)$, that represents a
    probability distribution over parameter space $\pmb{\theta}$, and a centroid
    $\bar{\pmb{f}_k^j} = \text{ff}_k(l(\arg\max_\theta G_k^j(\pmb{\theta})))$, where we use $l(\pmb{\theta})$ as a shorthand for the combination of \acrlong{CFM} and recursive application of the execution loop (see Figure \ref{fig:exploration_overview}).
\end{definition*}
Using generative models has the advantage that models from different clusters can be combined and jointly optimized to find a parameter set $\pmb{\theta}$ that generates a capability lying in several intended clusters.
The clustering procedure is visualized in Figure~\ref{fig:clustering_overview}. 

\begin{figure}
    \captionsetup{type=figure}
    \includegraphics[width=\columnwidth]{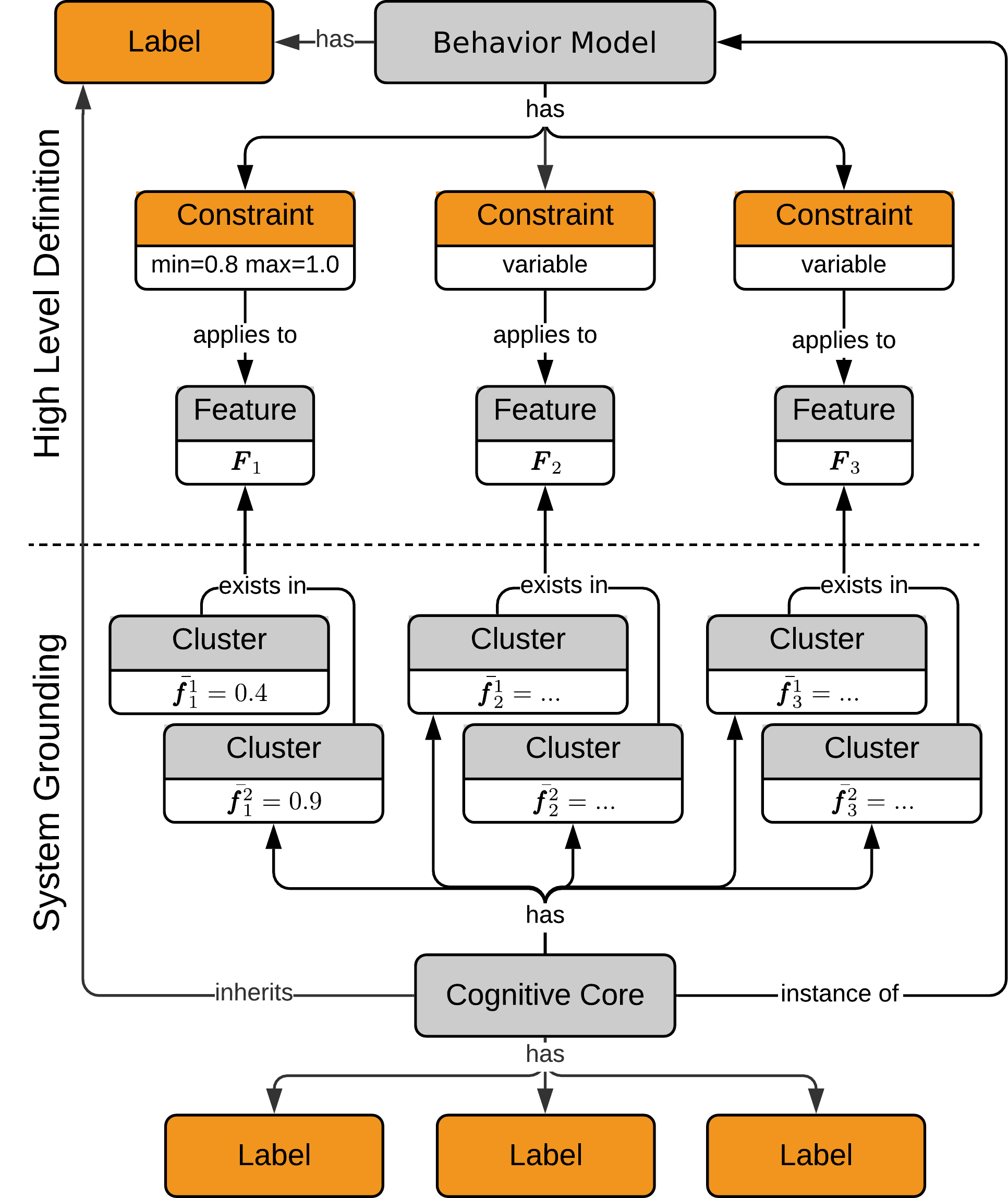}
    \caption{Relation between features, clusters, \acrlongpl{BM}, \acrlongpl{CC} and \acrlongpl{SA}. The \acrlong{BM} is defined by constraints in \acrlongpl{FS}. This behavior can be grounded as a \acrlong{CC}
    for a specific system when clusters for this system exist that fulfill 
    these constraints. Constraint fulfilling clusters are linked to the \acrlong{CC}. The \acrlong{CC} inherits the generic label of the \acrlong{BM}, but can have 
	more that describe specifics for this robot. The constituents of the \acrlong{SA} are colored orange.}
    \label{fig:cc_ccm}
\end{figure}

After clustering, \added{robot-specific} \acrlongpl{CC} can be instantiated. \Acrlongpl{CC} are defined as:
\begin{definition*}[Cognitive Core]
    A \emph{\acrlong{CC}} is an executable grounding of a \acrlong{BM} for a specific robotic system, where constraints of the \acrlong{BM} are checked against cluster centroids 
    $\bar{\pmb{f}_k^j}$. Clusters that satisfy these constraints are linked to the \acrlong{CC}. A \acrlong{CC} can only be generated when all \acrlong{BM}
    constraints can be met. 
\end{definition*}
These \acrlongpl{CC} are described by a \acrlong{SA}:
\begin{definition*}[Semantic Annotation]
    A tuple $\text{SA} =\left(L, \mathcal{X} \right)$, where $L$ is a set of labels, $|L| \geq 1$ and $\mathcal{X}$ is the set of constraints.
\end{definition*}
By default, the \acrlong{CC} inherits the labels and constraints from its
\acrlong{BM}, but the \acrlong{SA} can be augmented by \added{robot-specific}
information. This \acrlong{SA} is the main interface between the generation of
\acrlongpl{CC} and the reasoning processes
described in Section \ref{sec:reasoning}.
The relation between \acrlongpl{FS}, clusters, \acrlongpl{BM}, \acrlongpl{CC} and \acrlongpl{SA} is illustrated in Figure~\ref{fig:cc_ccm}. 

In this framework, the execution of a behavior on a specific robot, i.e., the
execution of a \acrlong{CC}, comes down to finding a parameter set
$\pmb{\theta}_{max}$ that jointly maximizes all generative cluster models
adhering to the constraints of the \acrlong{BM}. If a \acrlong{BM} includes
variable constraints, each target value $\pmb{f}_k^{tar}$ in the corresponding
\acrlong{FS} $\pmb{F}_k$ needs to be assigned. The \acrlong{CC} then finds the cluster models with closest centroids to the 
variable inputs. The \acrlong{CC} effectively uses a constraint checking
function $cc(c_k^j)$ to determine the relevant clusters, where
\begin{equation*}
	cc(c_k^j) = 
	\begin{cases}
	1  & \quad \text{if type is "min/max" and } \\
	   & \qquad \text{min } < \bar{f_k^j} < \text{ max}\\
	1  & \quad \text{if type is "variable" and } \\
	   & \qquad \bar{f_k^j} = \arg \min_{\bar{f_k^i}} m_k\left(\bar{f_k^i}, f_k^{tar}\right) \\
	0  & \quad \text{else. }
	\end{cases}
\end{equation*}
with function $m_k$ as the metric of the \acrlong{FS} $\pmb{F}_k$. 
Note that this implies that several cluster models in the same \acrlong{FS} can fulfill a min/max constraint.
The product of all currently relevant cluster models, \added{i.e.,} the models $G_k^j$ for which $cc(c_k^j)=1$, results in a new probability distribution.
The maximum of this distribution corresponds to a parameter set $\pmb{\theta}_{max}$ that has the highest likelihood of generating a capability that lies within all relevant clusters when used as input to the \acrlong{CFM} and the execution loop (see Figure \ref{fig:exploration_overview}).
The maximization step is then formally written as:
\begin{align*}
    & \pmb{\theta}_{max} = \arg\max_{\theta} \prod_{\pmb{M}} G_k^j(\theta),
\end{align*}
where $(k,j) \in \pmb{M}$ if  $cc(c_k^j) = 1$.
Since this approach is based on probabilistic modeling, it is possible that the
\acrlong{CAP} associated with $\pmb{\theta}_{max}$ violates a constraint.
However, assuming a smooth mapping  $\pmb{\Theta} \to \pmb{F}_k$ via the
\acrlong{FF}s $\text{ff}_k$, the violation is likely mild. If not violating a particular constraint is important, e.g., to avoid collisions, different weights can be assigned to different constraints, which control the relative influence of the corresponding cluster models. Note that it is also possible that cluster models are combined that have close to or completely disjunct distributions. Thus, in practice a probability boundary has to be set under which the  maximization result $\pmb{\theta}_{max}$ is 
rejected and it is assumed that no capability exists that fulfills all constraints.

One important challenge of the approach is how behaviors are cast into the constraint-based, phenomenological \acrlong{BM} we use. Since we aim at semantics which 
is intuitively understandable, we rely on human interaction. Thus, the first option \pname{Q-Rock} provides is hand-crafting \acrlongpl{BM}. Although 
it requires some domain knowledge, this approach scales well in the sense that
once defined, the \acrlong{BM} can be grounded for many different robotic
systems. In addition, we also envision \added{semi-automated} approaches: (1) \Acrlong{BM}ling from observation of human examples, and (2) Modelling human evaluation functions with respect to a specific behavior.
Approach (1) is based on research on end effector velocity characteristics for deliberate human movement \cite{gutzeit2016automatic,gutzeit2019automated}. These movement characteristics can be formulated as \acrlong{FS} constraints and thus used 
to define \acrlongpl{BM}. For approach (2), it was shown that implicit bio-signals of the human brain and explicit evaluation of a human observing simulated robot behavior 
can be used to effectively train a model of the underlying evaluation function \cite{leohold2019active}, and to guide a 
robotic learning agent \cite{kim2017intrinsic}. Also here, \acrlong{FS} constraints can be derived from the trained evaluation function 
approximator and used to define the \acrlong{BM}.

At this point, we want to stress again that human interaction is absolutely
necessary in the \pname{Q-Rock} philosophy to define meaningful behavior.
Throughout this workflow step, human labelling is required for \acrlongpl{FS},
\acrlongpl{CC} and \acrlongpl{BM}. The robot itself, after exploration, has no
notion of causality, i.e., reaction to the environment, or purpose in what it is
doing. Thus it is not behaving in the actual sense. Only through human semantic
descriptions, \added{i.e.,} what it would look like if the robot would behave in a certain way, are the \acrlongpl{CAP} of the robot 
in the environment ascribed to a meaningful behavior. Once the \pname{Q-Rock}
database grows, we will explore automatically generated labelling of \acrlongpl{FS} based on similarity 
to already labelled ones, which could speed up the labelling process by providing reasonable first guesses.

To summarize, \acrlongpl{CC} derived from \acrlongpl{BM} are central entities in the \pname{Q-Rock} workflow, since they cast explored \acrlongpl{CAP} in a semantically 
meaningful form and provide a way to generate new \acrlongpl{CAP} that adhere to characteristics found by clustering. In addition, their \acrlong{SA} 
provides the basis for reasoning about the connection of possible robot behavior to the underlying hard- and software, which 
will be elaborated in the following section.

    \subsection{Reasoning}
    \label{sec:reasoning}
    Structural reasoning serves two purposes in \pname{Q-Rock}: 
(1) to suggest suitable hardware to solve a user-defined problem, and 
(2) to map an assembly of hardware and software to its function.
The former does not involve any type of active usage of the hardware and
software assembly, but exploits
knowledge about the physical structure, interface types and known
limitations / constraints when combining components, as well as their relation
to labelled \acrlongpl{CC}.

Essentially, structural reasoning establishes a bi-di\-rectio\-nal mapping between
assemblies of hardware and software components and its function.
Note, that we explicitly do not use the term robot here, since the result of the
mapping from capabilities might not be a single robot, but a list of hardware and
software components.

\subsubsection{Related Work}
\gls{KRR} is considered a mature field of research,
but there is still a gap between available encyclopedic knowledge and robotics.
\pname{KnowRob} \cite{tenorth2009knowrob}, as knowledge processing framework,
intends to close this gap and
provides robots with the required information to perform their tasks.
It builds on top of knowledge representation research,
making the necessary adaptations to fit the robotics domain where typically
much more detailed action models are needed.
The core idea behind \pname{KnowRob} is to automatically adjust the execution of a
robotic system to a particular situation from generic meta action models.
The platform is validated with real robots acting in a kitchen environment
with a strong focus on manipulation and perception.
Beetz et al. combine \pname{KnowRob} with the usage of CRAM \cite{Beetz:2010:CRAM},
which serves as a flexible description language for manipulation activities.
CRAM in turn is used with the \gls{SDL} which
links capabilities with abstract hardware and software requirements through an ontological
model. As a result, symbolic expressions in CRAM can be grounded depending on
the available hardware.
CRAM is, however, not a planning system that can be used to solve arbitrary
problems. Instead it can formulate a plan template for an already solved planning problem.

Meanwhile, reasoning in \pname{Q-Rock} aims at using planning, in particular
\glspl{HTN}, to generically formalize a problem in the
robotic domain and to generate an action recipe as solution.
\gls{HTN} planning is an established technology with a number of
available planners such as CHIMP~\cite{stock2015online},
PANDA~\cite{Bercher:2014:PANDA} or SHOP2~\cite{Nau:2003:SHOP2}, but
there is still no de-facto standard language comparable
to the \gls{PDDL}~\cite{Fox:2003:PDDL} in the classical planning domain.
H\"oller et al.~\cite{hoeller:2020:hddl} suggest an extension to \gls{PDDL}
hierarchical planning problems named \gls{HDDL} to address
this issue. Nevertheless formulating an integrated planning problem which
includes semantic information remains an open challenge.

\subsubsection{Approach \& Formalization}

\begin{figure*}
    \captionsetup{type=figure}
    \includegraphics[width=\linewidth]{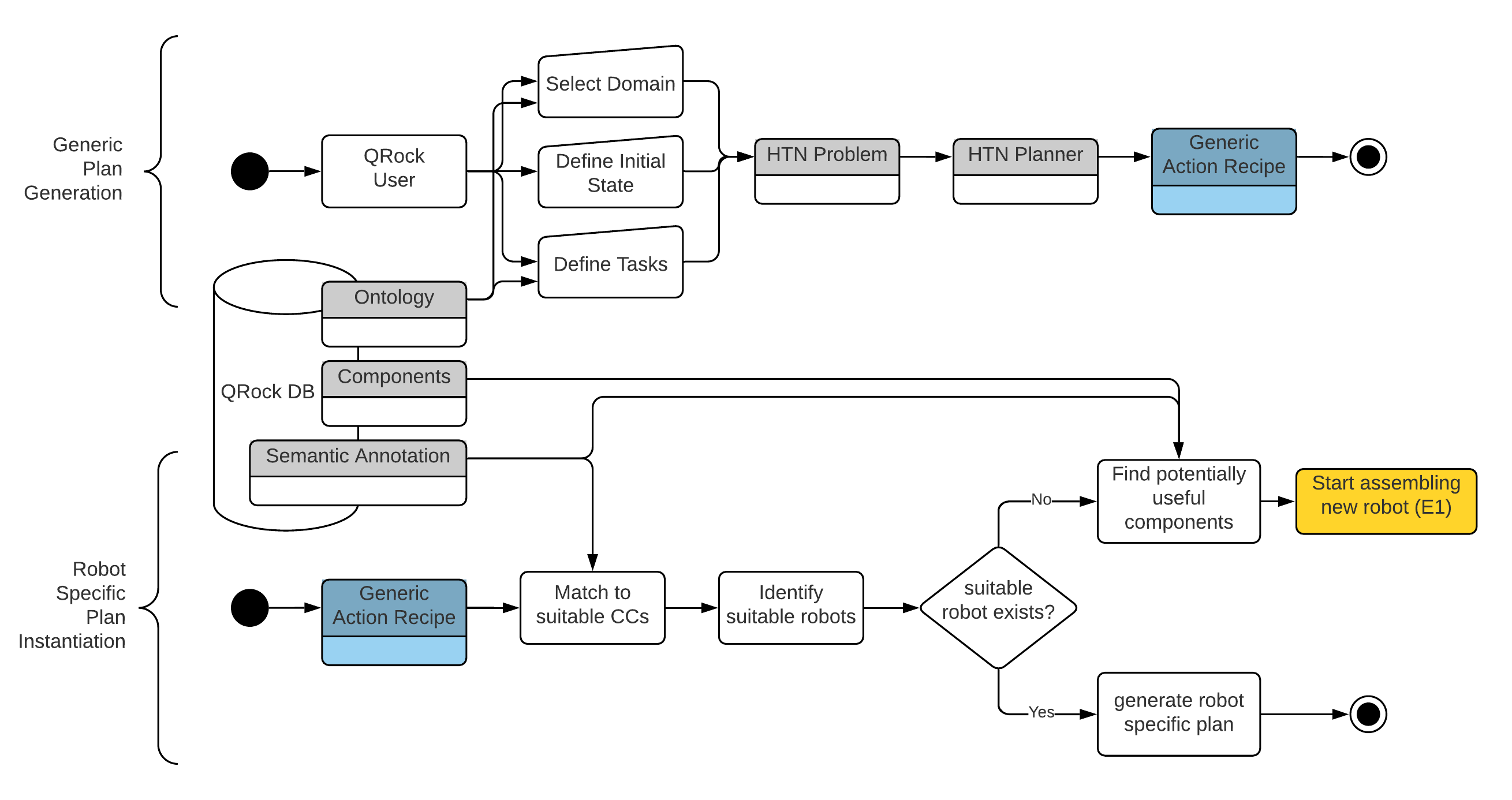}
    \caption{Outline of the top-down reasoning, which firstly involves the definition of
        a (planning) problem and the subsequent generation of a generic
    solution. Secondly, capable robots are identified to provide a robot
specific plan, or alternatively only to suggest components that might be relevant
to \added{design} a capable robot.}
    \label{fig:reasoning:top_down_workflow}
\end{figure*}

\paragraph*{Top-Down: Identification of capable systems}
\label{sec:reason:top_down}
We start by describing the process of structural reasoning from entry point E2
into the \pname{Q-Rock} development cycle (see Figure~\ref{fig:qrock_cycle}).
The workflow for the top-down reasoning is depicted in
Figure~\ref{fig:reasoning:top_down_workflow}.

To enter the cycle at E2
a user has to provide a description of an application problem to solve, i.e., defining
tasks that should performed and the application environment \added{including the
initial state.}
The problem is described with a general language and is firstly hardware agnostic. 
This means, neither does the application description explicitly state the use of a
particular robot nor a robot type.
While an input using natural language would be desirable for users to describe
their application,
\pname{Q-Rock} uses a planning language like \gls{PDDL} or as directly
machine readable format.
Formulating the application problem firstly generically and secondly as hierarchical planning problem allows the
decomposition into a sequence of atomic / primitive tasks, where $p \in P$ denotes a primitive
planning task and $P$ denotes the set of all primitive tasks.
\pname{Q-Rock} extends state-of-the-art 
planning approaches by (a) introducing a \acrlong{SA} for each primitive task, and
(b) representing the domain description, i.e., all tasks and decomposition
methods, with an ontology.

The \acrlong{SA} of a primitive task comprises a constrained-based
description of what a task does in the classical sense of planning effects,
i.e.,
what it requires to start the execution as preconditions and the condition
that have to prevail during an execution.
All conditions including pre/prevail and post can be tested upon using a predefined set of predicate
symbols, which describe the partial world state including environment state
$\pmb{s}^{obs}$ and robot state $\pmb{s}$.
Hence, the \acrlong{SA} of a primitive task also includes pre and prevail
conditions that link to the state of hardware and 
software components.

\begin{definition*}[Semantically Annotated Primitive Task]
    A \emph{semantically annotated primitive task} $\emph{p}^{+} \in P^{+}$ is a tuple of a
primitive planning task $p$ and a semantic annotation $SA$, so that $\emph{p}^{+} =
(p,SA)$. $P^{+}$ denotes the set of all semantically annotated primitive tasks.
\end{definition*}

The top-down reasoning is based on a predefined planning vocabulary
$\mathcal{V}_p = (P,C,\pmb{d},\emph{sa})$ to specify problems, 
here representing a particular planning domain description, where the vocabulary consists of 
primitive ($P$) and compound tasks ($C$), decomposition methods $\pmb{d}$ for
compound tasks, and a mapping function $\emph{sa}: P \to \mathcal{SA}$, 
$\mathcal{SA}$ denoting the set of all semantic annotations.
The top-down reasoning process is initially limited with respect to the
expressiveness of this application specification language.

Transforming the user's problem into a planning problem and solving it
results in a collection of plans, where each plan in this collection represents
a robot type agnostic solution.
This does not imply, however, that the requested task is solvable with current
available hardware.
Each semantically annotated primitive task that is part of a solution has requirements for its
execution including, but not limited to environmental, temporal, and hardware
and software constraints.
Therefore, an additional validation of these constraints has to be performed.

\begin{figure}
    \centering
    \includegraphics[width=.95\columnwidth]{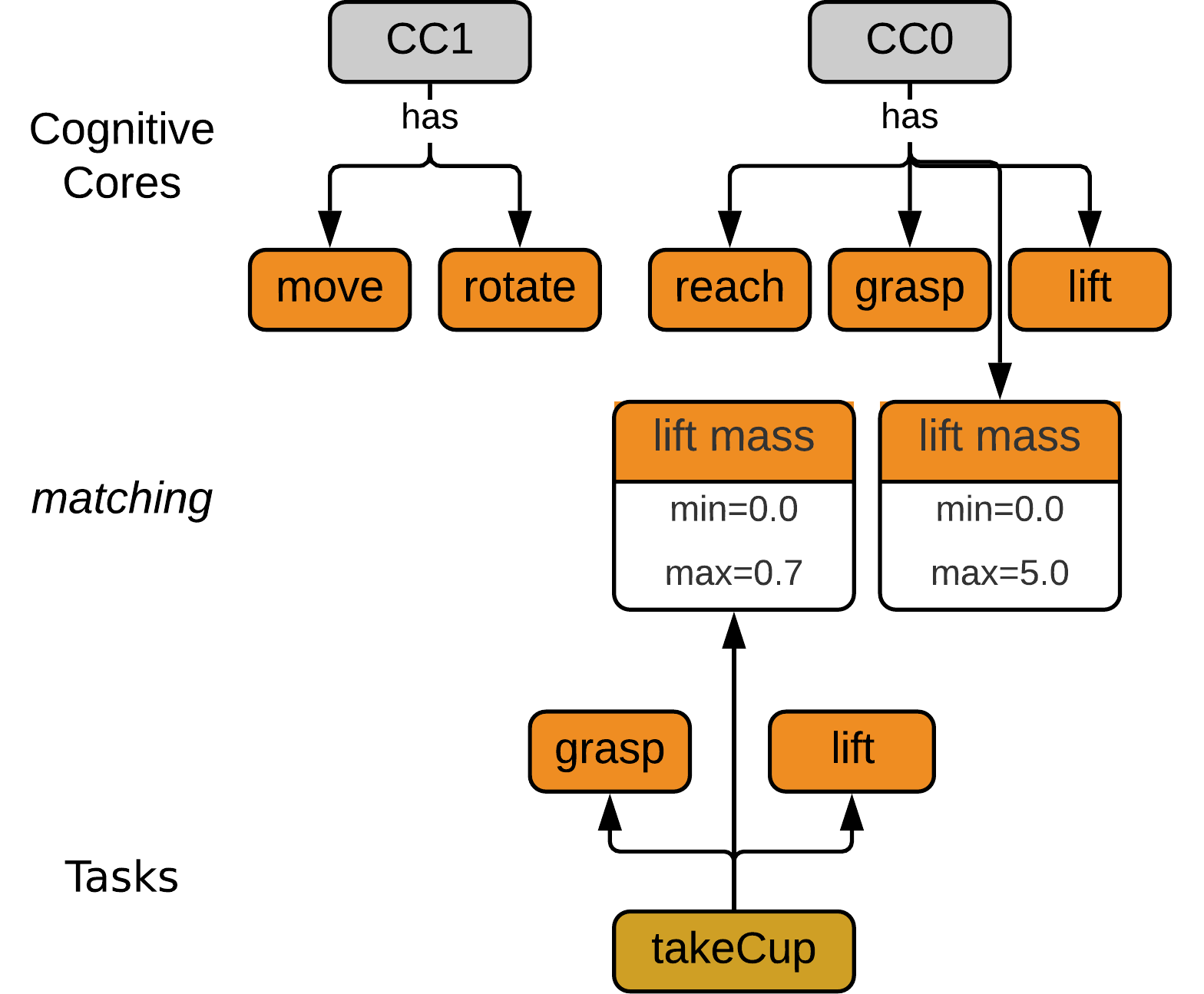}
    \caption{Matching of \acrlongpl{SA} in order to map from a task to a \acrlong{CC}
    that can perform this task}
    \label{fig:reasoning:semantic_annotation_as_glue}
\end{figure}

Requirements to execute a plan can be extracted from 
the \acrlongpl{SA} belonging to all of its semantically annotated primitive tasks,
in the simplest case by the use of labels which are organized in an ontology.
\Acrlongpl{SA} also describe \acrlongpl{CC} as 
explained in Section~\ref{sec:classify_annotate}.
Such a description might be incomplete in the sense that it does not catch every
detail of the behavior of a \acrlong{CC}, but it serves to outline the semantics in an
abstract and also machine processable way.
Furthermore, it allows to match \acrlongpl{SA} of tasks against \acrlongpl{SA}
of known \acrlongpl{CC}.
Thereby identifying \acrlongpl{CC} that can be used to
tackle the stated problem (see
Figure~\ref{fig:reasoning:semantic_annotation_as_glue}).
Each \acrlong{CC} maps to a single robotic system, but primitive tasks can map
to different \acrlongpl{CC}. Finally, a solution is only valid if a single suitable system
which is capable to perform all tasks can be identified.
While this concept of matching tasks and \acrlongpl{CC} can also be used to map
to multiple systems that cooperate to solve the stated problem, \pname{Q-Rock}
focuses on single robots for now.

As outlined before, \pname{Q-Rock} aims at a planning approach which does not focus on a particular
robotic systems, but provides abstract solutions.
Although no specific robot types are considered, solutions still can comprise hardware
requirements to solve a particular task.
For instance, a requirement could be the emptiness of a gripper before starting a gripping activity.
This particular precondition, however, implies also the availability of a gripper and thus
restricts the applicable robot types that can be used to perform for this task to those
that have a gripper. A selected target object might induce additional constraints for
lifting mass or handling soft objects, so that only a particular type of gripper
can be used.

Effectively, the following structural requirements exist for hardware and software components:
\begin{inparaenum}
\item existence of hardware and software components in the system, and
\item particular (sub)structures formed by hardware and/or software components.
\end{inparaenum}
Additionally, functional requirements exist which might imply structural
requirements, so that functional requirements can be considered as higher-order
predicates for tasks. These could be implemented similar to using semantic
attachments for planning actions as suggested by \cite{Dornhege:2012:TFDM}.
Workspace dimensions and required maximum reach are examples of an extended
task description, which limits the range of systems applicable for
this task.

To create \acrlongpl{SA}, \pname{Q-Rock} uses a corresponding language
$\mathcal{L}$.
Meanwhile, \pname{Q-Rock} uses an ontology to represent the vocabulary
$\mathcal{V} \supset \mathcal{V}_p \cup \mathcal{V}_{SA}$ of this language,
which combines the planning vocabulary $\mathcal{V}_p$ and the semantic annotation
vocabulary $\mathcal{V}_{SA}$ which permits to specify components, labels (corresponding to behavior types) and constraints.
While labels allow to classify and categorize behaviors, constraints allow to
detail or rather narrow these behaviors further, e.g., manipulation with a
constraint to manipulate a minimum of 100\,kg mass poses a significant hardware
constraint.

\added{\Acrlongpl{SA}} characterize primitive tasks as well as \acrlongpl{BM}
and \acrlongpl{CC} and are therefore essential to link \pname{Q-Rock}'s
clustering step with top-down reasoning.

\paragraph*{Bottom-Up: From Structure to Function}\label{sec:reason:bottom_up}
While the top-down reasoning process tries to find suitable hardware for a given
task, the bottom-up approach aims at finding the functionality or rather tasks
that a component composition can perform.
The bottom-up identification of a robot's function is based on
a formalization introduced by Roehr~\cite{Roehr:2019:Thesis}
who establishes a so-called organization model to map between a composite system's functionality and the
structure of components.
Functionalities can be decomposed into their requirements on structural system
elements.
As a result, the known structural requirements for a functionality can be matched against (partial) structures of a
composite system to test whether this functionality is supported.
\begin{figure}
    \captionsetup{type=figure}
    \includegraphics[width=\columnwidth]{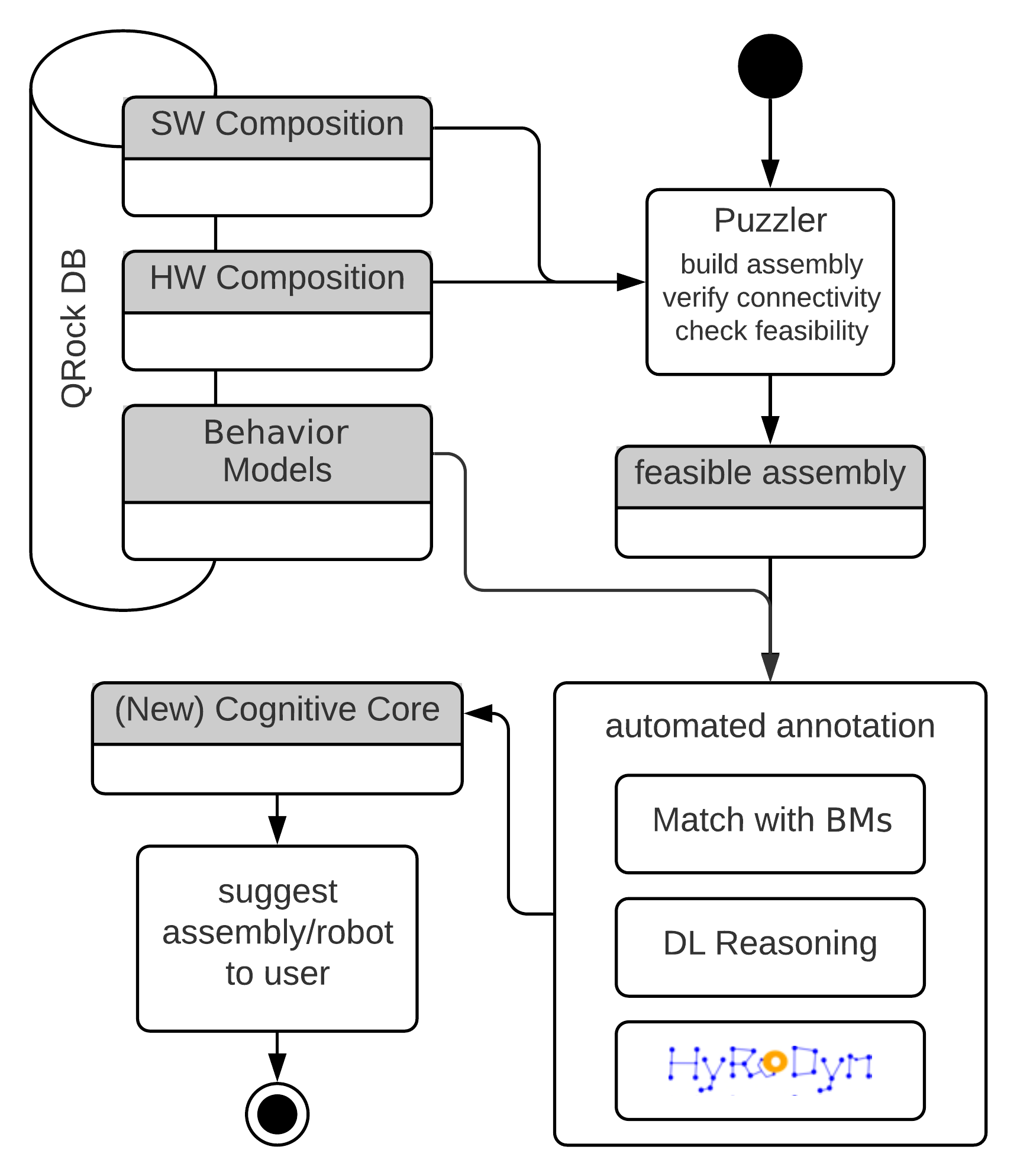}
    \caption{The bottom-up reasoning is based on a combinatorial,
        constraint-based and heuristic search approach in order to identify
        feasible assemblies, which will be subsequently characterized in an
    automated way.}
    \label{fig:reasoning:bottom_up_workflow}
\end{figure}
Figure~\ref{fig:reasoning:bottom_up_workflow} depicts the bottom-up reasoning
workflow, where an essential element of the bottom-up reasoning is the identification of
feasible composite systems or rather assemblies.
The combination of components requires knowledge about the interfaces of the components
and permits a connection between any two components only if their interfaces
are compatible.
Multiple interfaces
might be available for connection and physical as well as virtual (software)
interfaces can be considered.
Roehr~\cite{Roehr:2019:Thesis} limits the bottom-up reasoning to a
graph-theoretic approach while excluding restrictions arising from the actual
physical properties of the overall component, e.g., shape or mass.
\pname{Q-Rock} will remove that restriction and analyse the actual physical
combinations of components as part of the so-called \emph{puzzler} component.
\changed{E6}{The \emph{puzzler} component composes new assemblies from a known set of
    atomic components by creating links between compatible interfaces of atomic components.
    By using the existing D-Rock tooling, and extending the component model
    specification with ontological knowledge from \gls{Korcut}~\cite{Korcut:2021}, the newly defined assembly is
    loaded into blender and exported to \gls{URDF} along with additional other material and
    sensor information. Subsequently, the new assembly can be physically
    validated and explored in simulation.
    Based on the \gls{URDF} representation we will use
    \gls{HyRoDyn}~\cite{2019_HyRoDyn_IDETC,2020_HyRoDyn_JMR} for characterizing
    the robot, e.g., by computing basic properties of the
    robot in zero configuration, and analysing configuration space, work\-space and forces.
    Furthermore, the assemblies will be semantically
    annotated by (a) matching the structure to existing \acrlongpl{BM} and (b) reasoning on the ontological description.
}
An initial manual and later automated analysis of component structures in
existing robotic systems can serve as a basis to identify generic design patterns in robotics systems.
This can be used as heuristic to boost the bottom-up reasoning process.
The bottom-up reasoning process is triggered from new additions of software and
hardware to the database. Thereby, \added{augmenting} the \pname{Q-Rock} database by
adding new \acrlongpl{CC} \added{helps to increase} the options for solving user problems.

    \section{Results: A Use Case Scenario}
    \label{sec:use_case}
    \begin{figure*}[h!]
    \captionsetup{type=figure}
    \includegraphics[width=\linewidth]{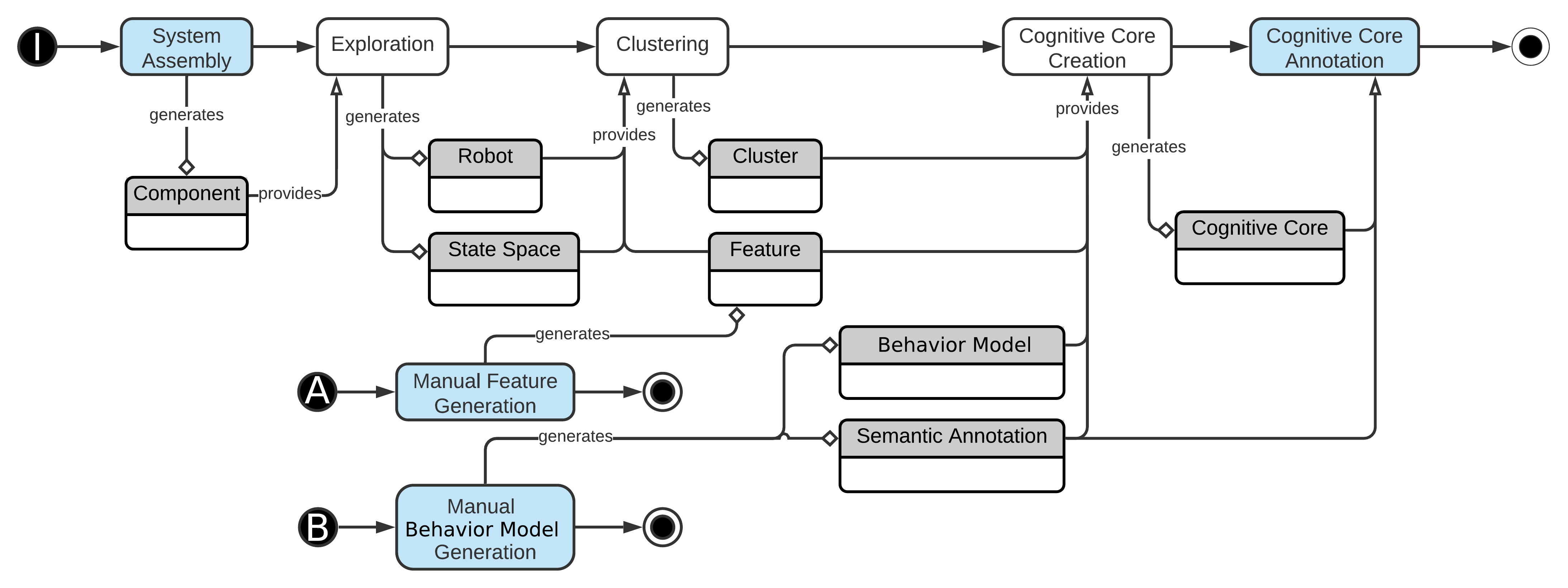}
    \caption{Workflow for our exemplary use case. Steps with required user interaction are shaded blue. 
    Relations between entities are avoided for clarity. The full entity relation diagram is visualized 
    in Figure~\ref{fig:entity_relationship_model}. We start with entry point A to manually define features. 
    Then proceed with entry point B and manually define a behavior model. Then
    the actual work flow starts at I. 
    A system is assembled using components from the database, generating a new component. This new component 
    is passed to the exploration step, which generates a robot and a state space entity in the database. Part of the 
    robot entity is the capability function, which is used in the clustering step, along with information about the 
    state space of the robot and the features in which to cluster. The clusters that are generated are used in 
    cognitive core creation, which grounds the previously defined behavior model for this robot. In the 
    cognitive core annotation step, the semantic annotation inherited from the
    behavior model is reviewed
    and possibly extended with robot-specific information by a human observer.}
    \label{fig:use_case_workflow}
\end{figure*}

\changed{R1.15}{The \pname{Q-Rock} development cycle combines
existing \acrshort{AI} technologies in order to
\begin{inparaenum}[(a)]
\item simplify the robot development process and
\item exploit the full capabilities of robotic hardware. 
\end{inparaenum}
In this section we provide a qualitative analysis of the \pname{Q-Rock} development
cycle on the basis of three complementary use cases which allow us to illustrate
the concept and locate conceptual as well as practical issues. The use cases
are:
\begin{inparaenum}[(1)]
\item system assembly, and then using the assembly for
\item solving a task and 
\item solving a mission.
\end{inparaenum}
For the system assembly the workflow starts at entry point E1 
(see Figure~\ref{fig:qrock_cycle}), namely constructing a robot model, exploring its movement capabilities, 
clustering these capabilities and generating a \acrlong{CC}.
The framework is designed to handle all kinds of capabilities that can be
described as trajectories in the combined robot and world state.
For this example and initial evaluation we focus, however, on movement capabilities.
}
\changed{R1.2}{We use three test systems - a 3-DOF robotic arm, a wheeled mobile base, and a combination of both - and create a \acrlong{CC} \emph{reach}, which
	moves the end effector from a given start configuration to a
	chosen target position in task space.}

\changed{R1.15}{
The schematic workflow is visualized in Figure~\ref{fig:use_case_workflow}, while Figure~\ref{fig:workflow:robot_configurator} highlights steps 
of this workflow which are triggered through our implemented website.
Several preparatory steps are, however, assumed and required for the workflow to run
including the definition of (a) \acrlongpl{FS} and (b) \acrlongpl{BM}.
We validate the creation of the \acrlong{CC} and integration of the
system, by retrieving the \acrlongpl{CC} again via its labels from the database.
}

\subsection{Preparation}
\changed{R1.15}{We predefine \acrlongpl{FS} which are
used in the clustering step and for defining a \acrlong{BM}. Note, that this is a
simplification which we will address in future implementation, e.g., by the development 
of automated feature learning approaches.
}

\paragraph*{Defining Feature Spaces}
To characterize a \emph{reach} movement the \acrlongpl{FS} firstly permit to extract the 
starting state and the end state of trajectories.
Further qualities of a trajectory such as its directness are
also included to penalize deviations of a trajectory
from the direct route.
As \acrlongpl{FS} we use: 
\begin{enumerate}
    \item $\pmb{F}_{start}$ with label 'start state' and the transformation
        function $f_{start}: L \to S$ which maps a given trajectory $l^{T}\in L$ to
        its start state $\pmb{s}^0 \in S$,
    \item $\pmb{F}_{end}$ with label 'end effector end state' and the transformation function 
    $f_{end}: E \to S$, which maps a given end effector state trajectory to its
    final state $\pmb{e}^T$,
\item $\pmb{F}_{dir}$ with label \changed{R1.8}{'end effector directness'} and the transformation 
    function $f_{dir}: L \to \mathbb{R}$ and \changed{R1.25}{
        \begin{multline*}
            f_{dir} = \frac{|s_e^0 - s_e^T|}{\sum_{i=0}^{T-1} |s_e^{i+1} - s_e^i|} \in (0,1]
        \end{multline*}
    }
\end{enumerate}
Note that the features $\pmb{F}_{end}$ and $\pmb{F}_{dir}$ use the end effector position,
which we assume is part of the world state $\pmb{S}^{obs}$, whereas $\pmb{F}_{start}$ operates on the internal state $\pmb{S}^{rob}$ of the robot.

\paragraph*{Defining Behavior Models}
A \acrlong{BM} provides the high level abstraction for a behavior,
thereby collecting the essential characteristics.
In the case of the \emph{reach} behavior the key characteristic is the directness,
which we expect to be high so that it is bound to a minimum and maximum degree.
Meanwhile the start and the end state are variable, since the \emph{reach} behavior
needs to be applicable in a range of situations with different target poses.
Hence, start and end state can be viewed as general input parameters to the
\acrlong{BM}.
\begin{align*}
    \mathrm{label}: & \quad \mathrm{reach} \\
    \mathrm{constraints}: & \quad \pmb{F}_{dir}: \mathrm{min:} \, 0.8 \quad \mathrm{max:} 
    \, 1.0 \\
    & \quad \pmb{F}_{start}: \mathrm{variable} \\
    & \quad \pmb{F}_{end}: \mathrm{variable}
\end{align*}
All defined \acrlongpl{FS}, the \acrlong{BM} and the \acrlong{SA} of the
\acrlong{BM},
which contains labels and constraints, are stored in the database to be
accessible for all development steps. 

Note that defining the \acrlong{BM} is an essential, but currently also a limiting
requirement, since the exploration of behaviors can only cover these predefined models.

\subsection{System Assembly}
To start the \pname{Q-Rock} development cycle at entry point E1 a robotic system is
required.
Predefined robots or rather existing assemblies can be used
to start the exploration.
One of the major motivations of the \pname{Q-Rock} development
cycle is, however, the capability to explore any kind of hardware designs /
assemblies and thereby support an open robot design process.

\begin{figure*}
    \centering
    \captionsetup{type=figure}
    \begin{subfigure}[t]{.45\textwidth}
        \includegraphics[width=\textwidth]{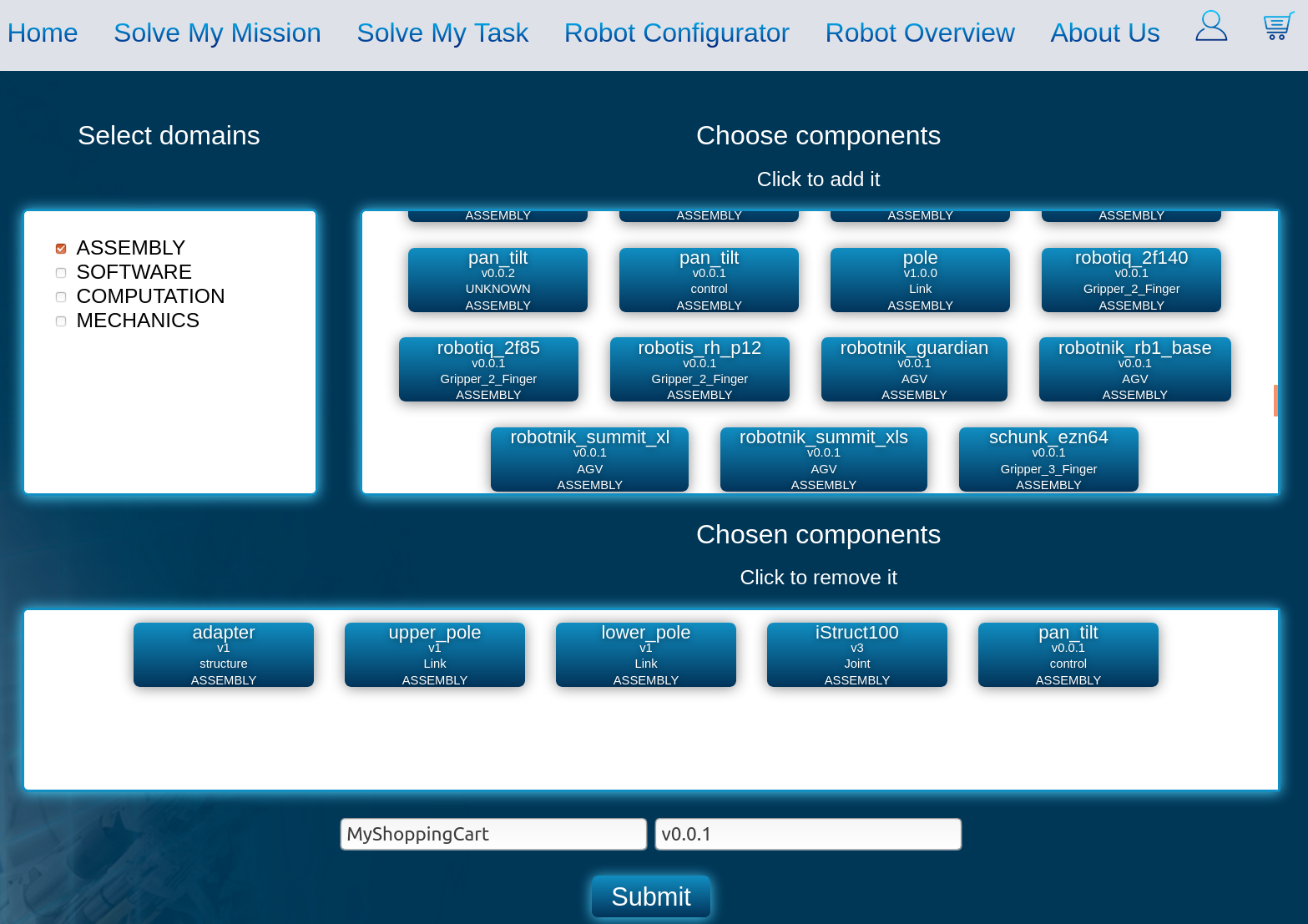}
        \caption{Choosing the components}
        \label{fig:workflow:choosing_components}
    \end{subfigure}
    \quad
    \begin{subfigure}[t]{.45\textwidth}
        \includegraphics[width=\textwidth]{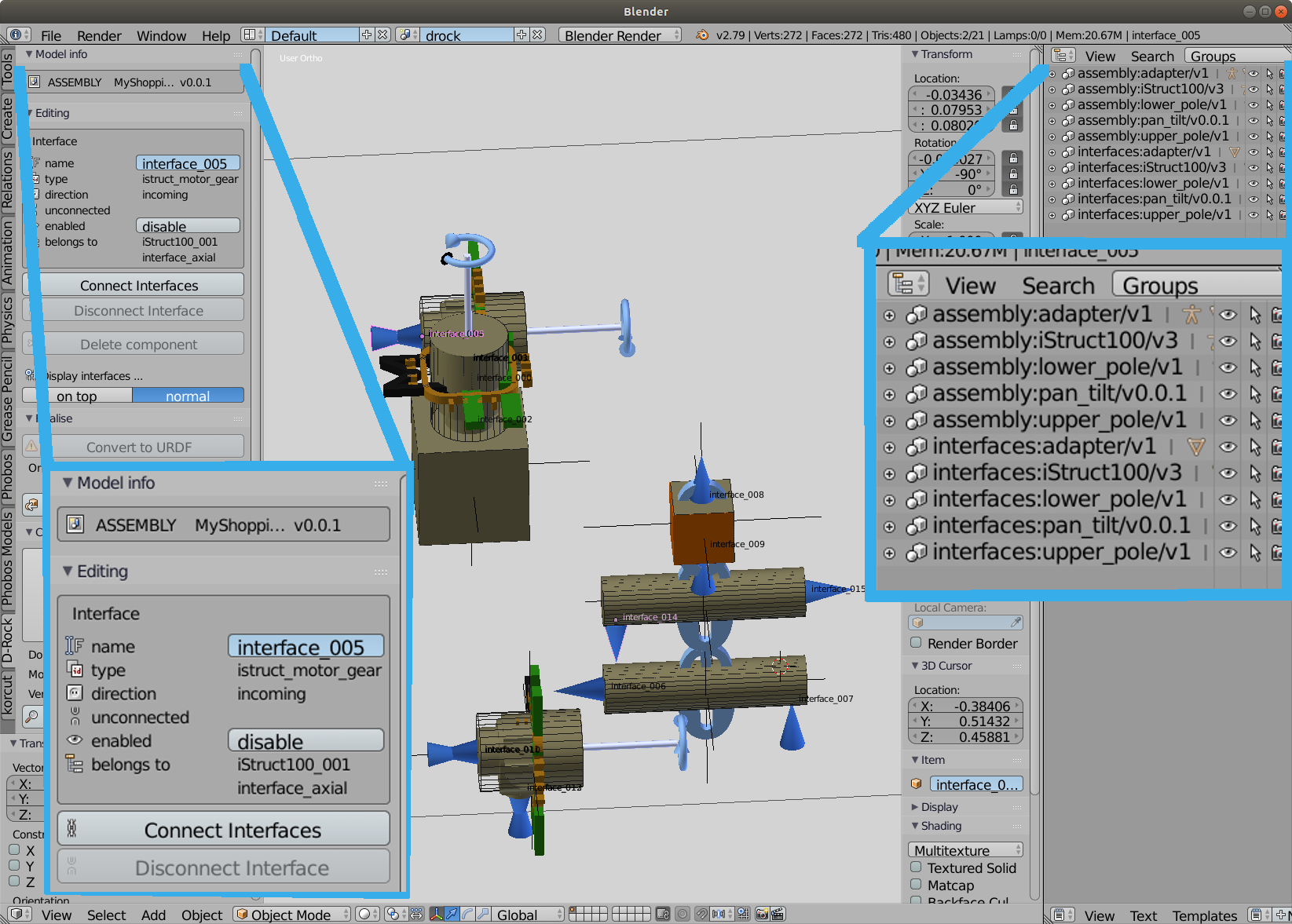}
        \caption{Assembling components with the help of Blender and Phobos}
        \label{fig:workflow:assembling}
    \end{subfigure}
    \begin{subfigure}[t]{.45\textwidth}
        \includegraphics[width=\textwidth]{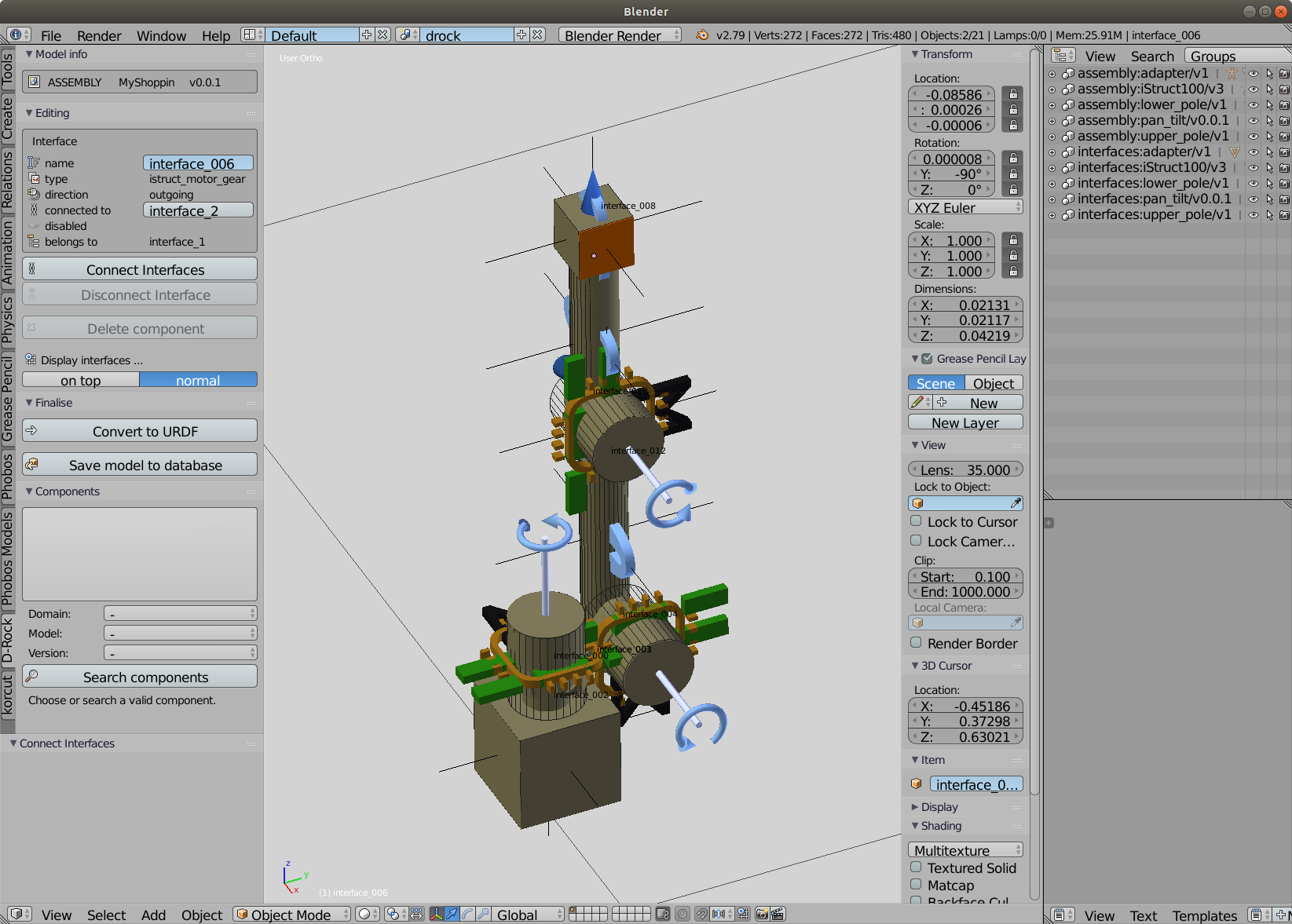}
        \caption{Saving the final assembly into the database}
        \label{fig:workflow:save_assembly}
    \end{subfigure}
    \quad
    \begin{subfigure}[t]{.45\textwidth}
        \includegraphics[width=\textwidth]{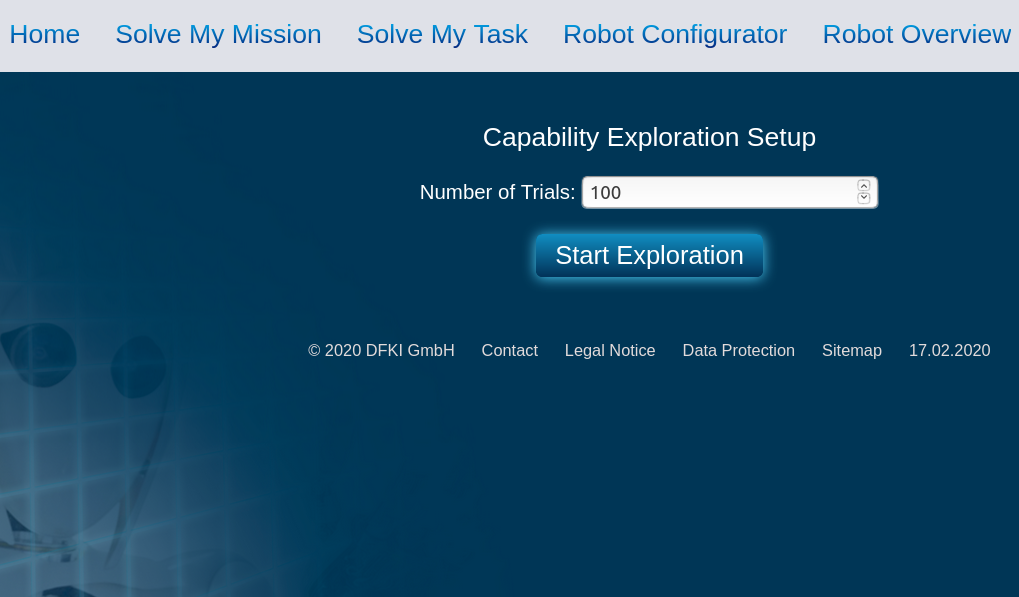}
        \caption{Parametrize the exploration}
        \label{fig:workflow:prepare_exploration}
    \end{subfigure}
    \begin{subfigure}[t]{.45\textwidth}
        \includegraphics[width=\textwidth]{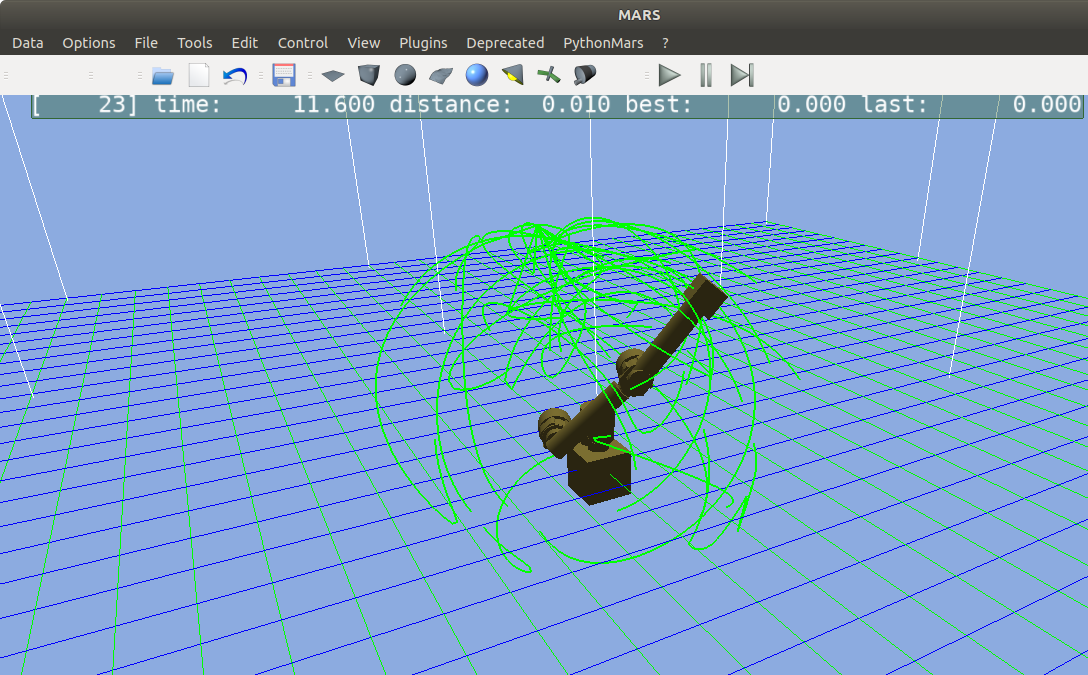}
        \caption{Explore the capability of the new assembly}
        \label{fig:workflow:explore}
    \end{subfigure}
    \quad
    \begin{subfigure}[t]{.45\textwidth}
        \includegraphics[width=\textwidth]{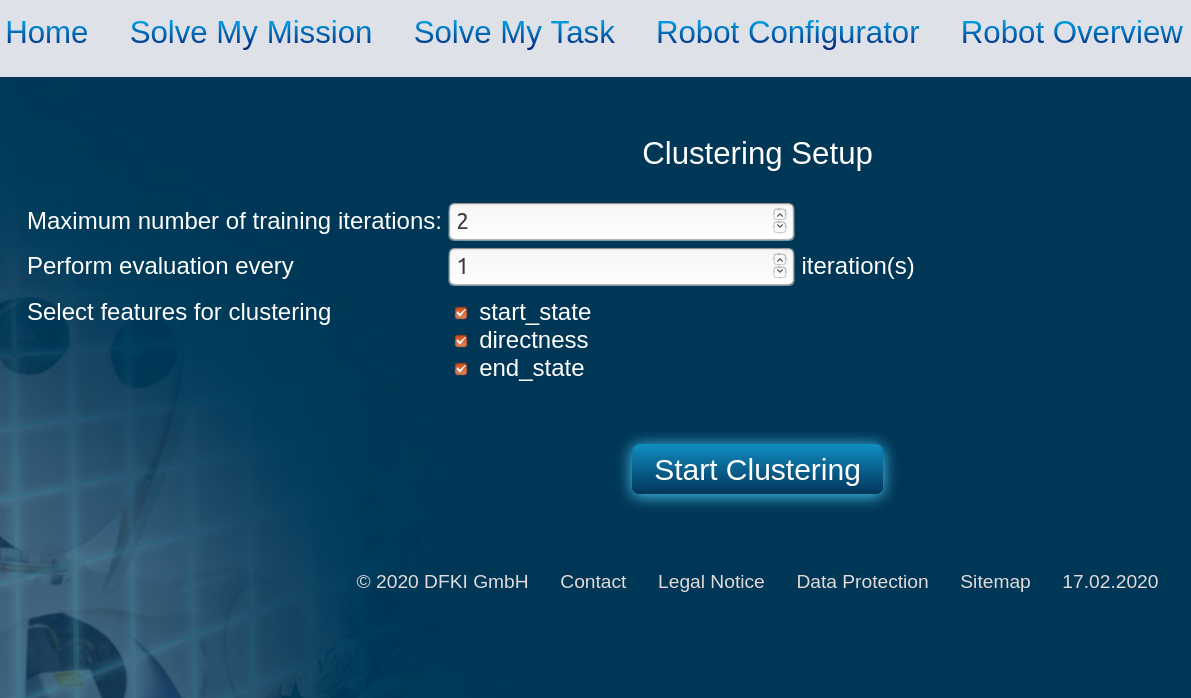}
        \caption{Clustering is applied and \acrlongpl{CC} identified from the
        existing \acrlong{BM} labelled \emph{reach}}
        \label{fig:workflow:clustering}
    \end{subfigure}
    \begin{subfigure}[t]{.45\textwidth}
        \includegraphics[width=\textwidth]{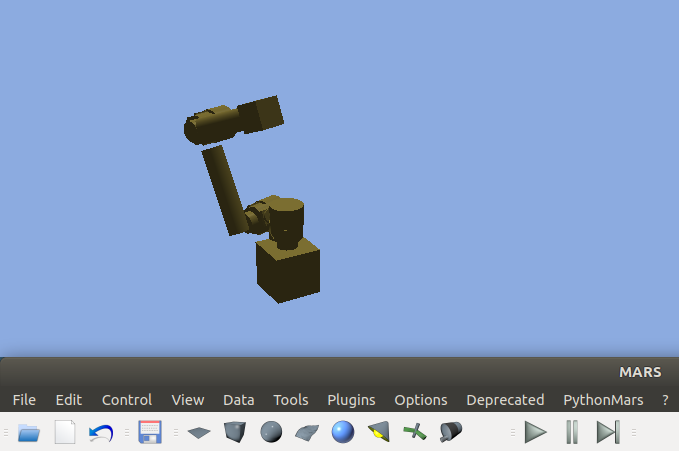}
        \caption{A video of the identified cluster performance is automatically rendered}
        \label{fig:workflow:video_rendering}
    \end{subfigure}
    \quad
    \begin{subfigure}[t]{.45\textwidth}
        \includegraphics[width=\textwidth]{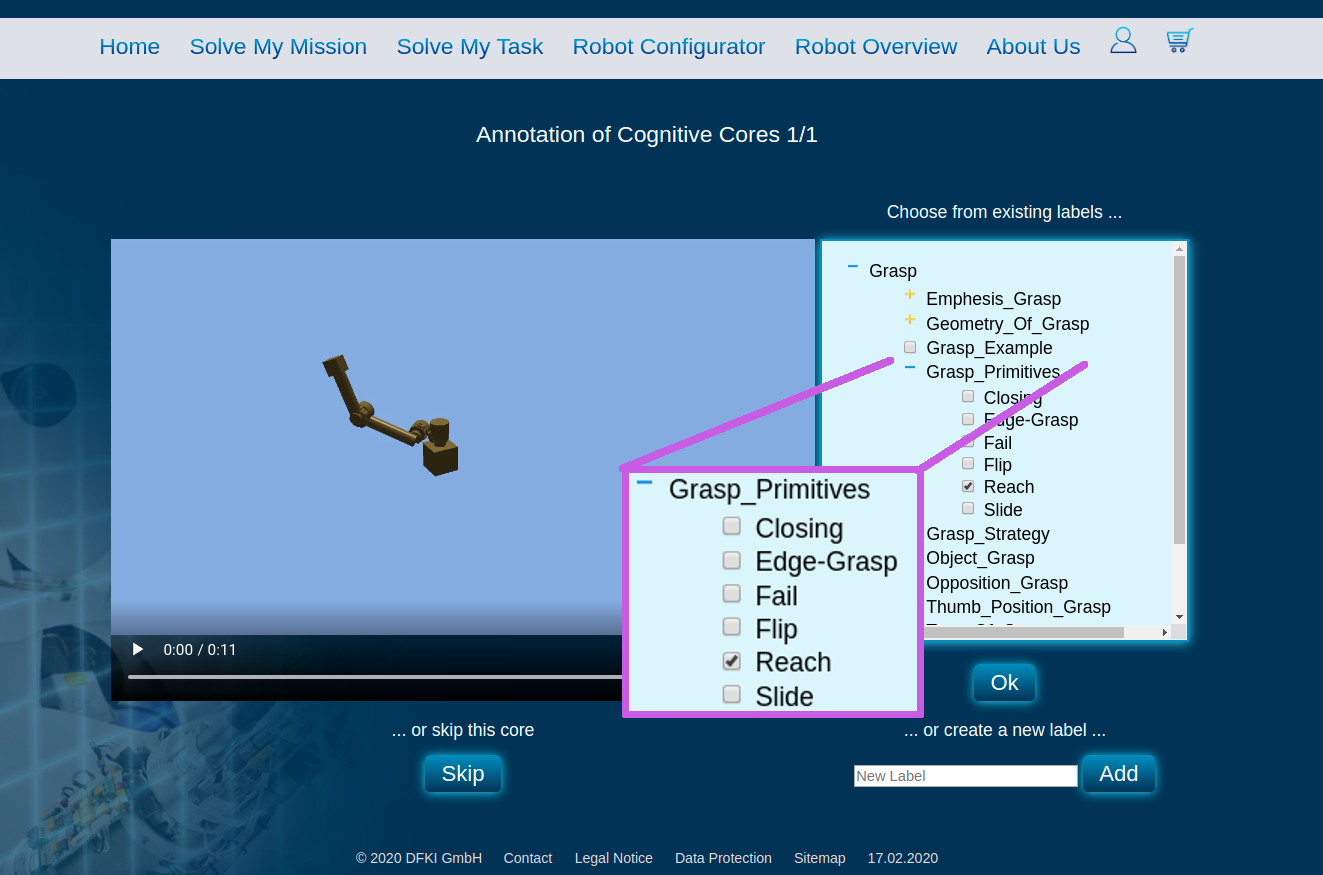}
        \caption{A user can watch videos of a \acrlong{CC} and annotate}
        \label{fig:workflow:annotate}
    \end{subfigure}
    \caption{\changed{R1.26}{The \emph{Robot Configurator} workflow takes advantage of
    exploration and clustering and allows to construct a robot first, which will
then automatically be explored and annotated.}}
    \label{fig:workflow:robot_configurator}
\end{figure*}

The so-called \emph{Robot Configurator} workflow permits a user to create a
robotic system in a simplified way, by combining a set of components that are
defined in the database.
\changed{R1.26}{Figure~\ref{fig:workflow:robot_configurator}} illustrates the steps.
Firstly, a user selects the desired items which are needed to build the robot and puts
them in a shopping basket (see Figure~\ref{fig:workflow:choosing_components}).
For the \emph{reach} example an assembly is built from the following items:
\begin{inparaenum}
    \item pan tilt unit,
    \item lower pole,
    \item joint motor,
    \item upper pole, and an
    \item end effector.
\end{inparaenum}
After the selection has been completed, a \acrshort{CAD} editor is started with the selected
items being already loaded (see Figure~\ref{fig:workflow:assembling}).
We use the open source editor Blender\footnote{http://www.blender.org} in combination
with the extended functionality of the Phobos plugin~\cite{Szadkowski:2020:Phobos} and
another \added{custom} plugin to interface with the database.
The user can build the desired system by selecting interfaces in the \acrshort{GUI} and
request to connect components through theses interfaces, which is only possible
if the selected interfaces are
compatible. \changed{E3}{Component interfaces and their compatibility are defined in hand curated ontologies.}
The overall procedure requires only very limited editing competencies of a user,
thus significantly lowering the entry barrier for physically designing a robot.
Once the final system has been assembled, the user can save the new design to
the database (see Figure~\ref{fig:workflow:save_assembly}).

\paragraph*{Exploration}
For the exploration of movement capabilities we chose a \acrlong{CFM} where the
parameters $\pmb{\theta}$ correspond \changed{R1.9}{to polynomial parameters}. The parameters
define an intended joint trajectory for all joints of the robot.
\changed{R1.9}{For a single joint the trajectory is defined by:
\begin{align*}
 q(\phi) = -\theta_0 (\phi - 1) + \theta_1 \phi + \sum_{i=2}^{N_{\theta} - 1}\theta_i \left( \phi^{i-1} - 1 \right)\phi
\end{align*}
where $\phi = \frac{t}{T}$ is a phase and $T$ the length of the trajectory. The
first parameter $\theta_0$ corresponds to the start position and $\theta_1$ to
the final position. The number of parameters per joint was set to $N_{\theta}=5$.}

\begin{figure*}
	\centering
	\captionsetup{type=figure}
	\begin{subfigure}[t]{.45\textwidth}
		\includegraphics[width=\textwidth]{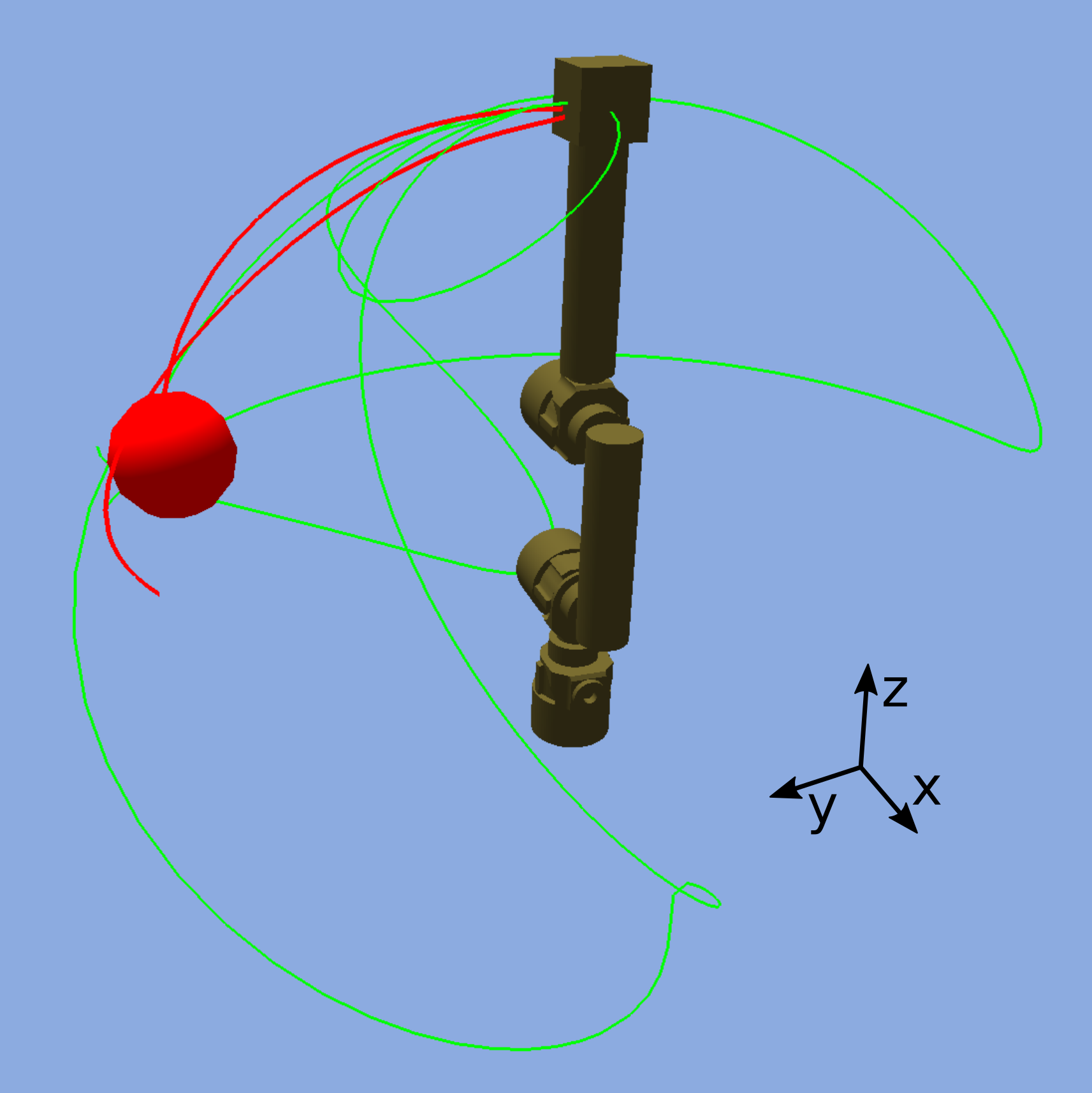}
	\end{subfigure}
	\quad
	\begin{subfigure}[t]{.45\textwidth}
		\includegraphics[width=\textwidth]{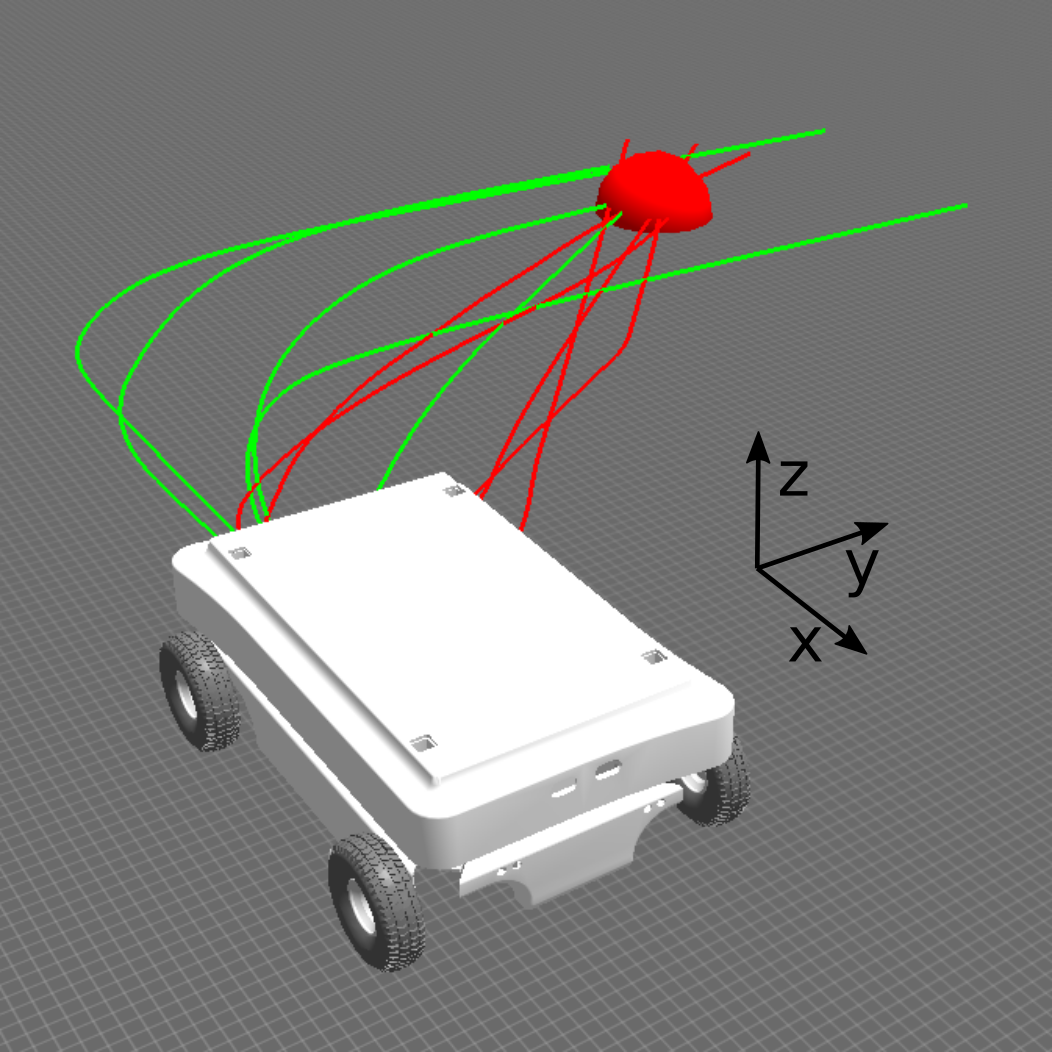}
	\end{subfigure}
        \caption{\changed{R1.2}{\emph{Reach} \acrlong{CC} for two robotic systems.
        Left: End effector trajectories generated by sampling from the
        \emph{reach} core of a 3-DOF arm (red lines) are shown with the requested target
area of the behavior (red ball) at $(x,y,z) = (0.1,0.3,0.2)$\,m. For comparison,
samples from the unconstrained \emph{reach} cognitive core - without the constraint on directness - are plotted in green. Right: Same as left, but for a wheeled mobile base, with a target
location at $(x,y,z) = (-1,1,0)$\,m.}}
	\label{fig:cc_reach_arm_base}
\end{figure*}

\changed{R1.9}{The \pname{Q-Rock} development cycle is also compatible with
    other motion representations such as splines and dynamic movement
    primitives. Meanwhile, this polynomial representation has the advantage that higher order
parameters only contribute if their value is different from zero. This
potentially eases the clustering process on this parameter space and it makes combining parameter sets with different $N_{\theta}$ straightforward.}

The intended
joint trajectories are passed to a controller which returns the robot
specific actions that are necessary to follow the desired trajectory as closely
as possible. The controller, where the desired states are specified by a
trajectory, takes the role of the \acrlong{CF} $\emph{cap}$ in this context. The
actions applied to the real system will finally result in a capability.

This \acrlong{CFM} ensures that the mapping from $\pmb{\Theta}$ to the capabilities is locally smooth,
\added{i.e.,} small changes \added{of} the parameters will lead to small changes in the corresponding trajectory. This is an 
important property for modelling clusters of trajectories in the Classification and Annotation process
(Section \ref{sec:classify_annotate}).

\begin{figure}
	\captionsetup{type=figure}
	\includegraphics[width=\columnwidth]{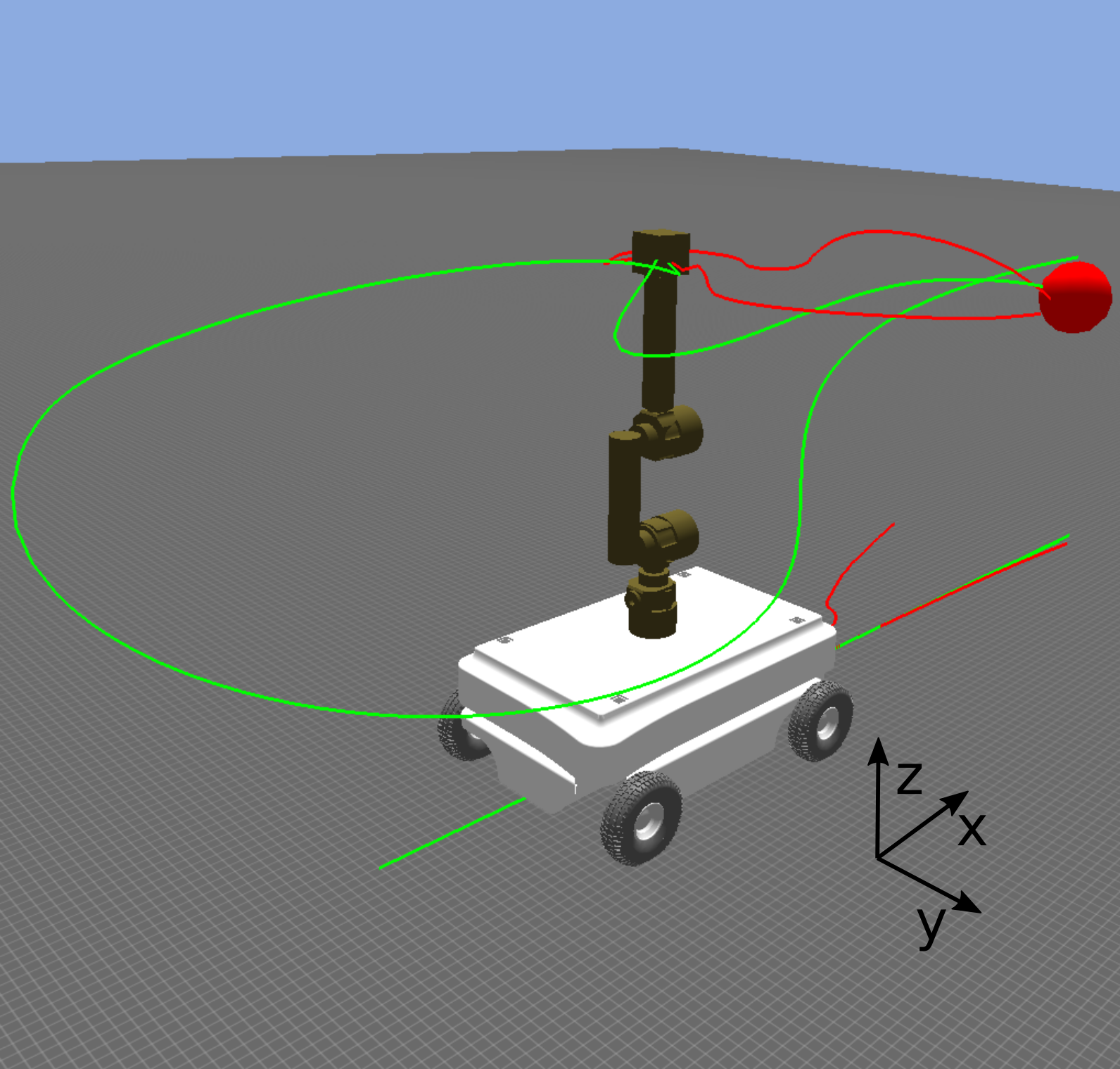}
        \caption{\changed{R1.2}{\emph{Reach} \acrlong{CC} samples for a combined system of 3-DOF arm 
	and mobile base. The red lines visualize the end effector motion of the
        arm of the combined system, thus including the motion of the
        mobile base. The target location is marked by the red ball at $(x,y,z) =
        (1.3,-0.1,0.5)$\,m, which is out of range of the arm alone, would the
        mobile base not move as well. The arm has a length of $\approx$ 0.4\,m. Note 
	that the directness constraint only applies to the end effector motion, which is the end 
	effector of the arm for this combined system, and not the task space
motion of the base. For comparison, samples from an unconstrained \emph{reach}
cognitive core are plotted in green. As expected, samples from this unconstrained core reach the target location more indirectly.}}
	\label{fig:reach_unconst_mob_base}
\end{figure}

To test the exploration approach, as described in Section~\ref{sec:exploration}, $10^6$ trajectories 
were generated for a robotic 3-DOF arm. Each trajectory has a length of $T = 4\,\text{s}$.
Five parameters specify the motion for every degree of
freedom, i.e., each parameter lies within $[-\pi,\pi]$, resulting in 15 parameters for the whole 3-DOF robot. \changed{R1.27}{The parameter space has been sampled uniformly.}. \changed{R1.2}{The same procedure 
was applied to a mobile base (see Figure \ref{fig:cc_reach_arm_base} right) and a 
combined system of mobile base and 3-DOF arm (see Figure \ref{fig:reach_unconst_mob_base} right). For the mobile base, 
the power series trajectories were applied to the velocity of the four wheels with a parameter 
range of $[-2\pi,2\pi]$.
\added{Three p}arameters were used to parametrize the motion, leading to 12
\added{p}arameters in total. The trajectory length was 
set to $T = 20\,\text{s}$. The combined system thus has a total of 27 parameters.}
The validation model has been implemented as a fully connected neural network
with four hidden layers and 22,102 parameters in total. Training for the 3-DOF arm was performed for 
20 training steps with a batch size of 100. 
After training, the validation model had an accuracy of
    96.6 $\pm$ 0.4 \% on unseen data \changed{R1.10}{of $10^4$ trajectories} for predicting whether a parameter set
corresponds to a valid motion.

\begin{figure*}
	\centering
	\captionsetup{type=figure}
	\begin{subfigure}[t]{.45\textwidth}
		\includegraphics[width=\textwidth]{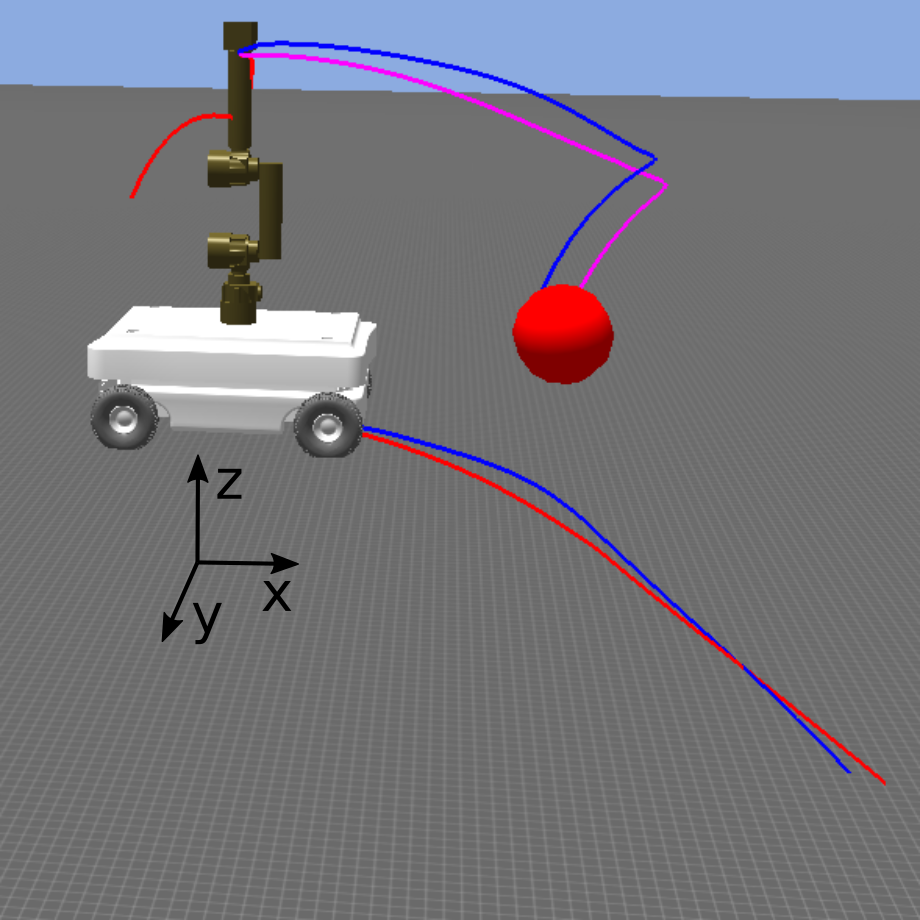}
	\end{subfigure}
	\quad
	\begin{subfigure}[t]{.45\textwidth}
		\includegraphics[width=\textwidth]{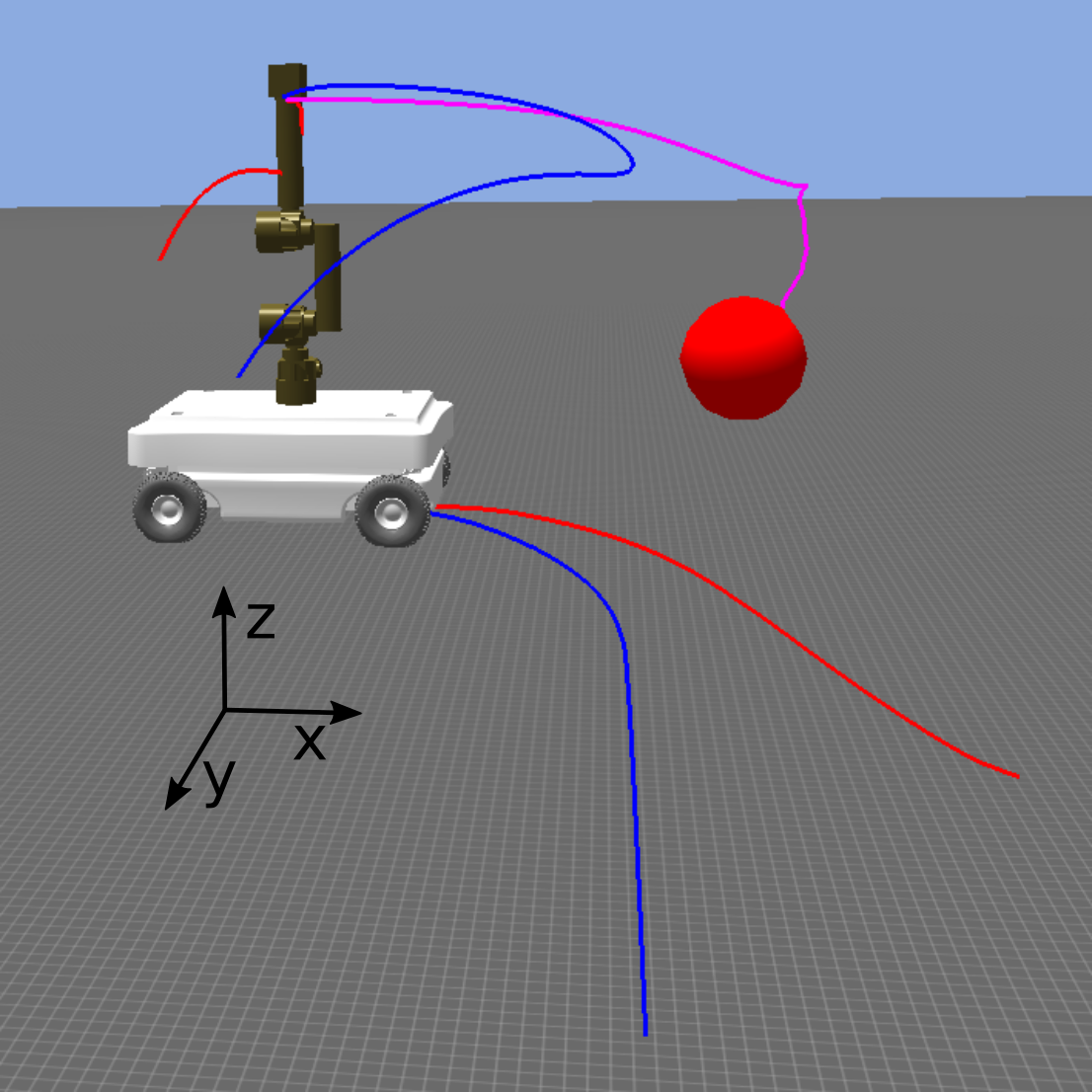}
	\end{subfigure}
	\caption{\changed{R1.2}{Parallel execution of \emph{reach} \acrlongpl{CC} of subsystems. The \acrlongpl{CC}
                were trained on each system individually and are samples of the same \acrlongpl{CC} shown in Figure \ref{fig:cc_reach_arm_base}. Left: The red lines show the end effector motion of each system in isolation, the magenta line the expected motion of the arm end effector of the combined system under the assumption that each subsystem behaves as in isolation. The red ball marks the requested target for the combined motion at $(x,y,z) = (1,2,0.5)$ m, outside of the task space of the arm alone, which has a length of $\approx$ $0.4$ m. The actual motion of the base and the arm of the combined system is shown by the blue lines. Right: Same as left, but for a different target position $(x,y,z) = (1.2,1.7,0.5)$ and different samples from the \acrlongpl{CC} for each subsystem, in which the dynamics of the subsystems lead 
	to a non-negligible interaction.}}
	\label{fig:cc_combined_reach}
\end{figure*}

\paragraph*{Clustering}
For clustering, the same capability set as for training the validation model was used. Without loss of generality, we implemented standard k-means
clustering as the clustering strategy \cite{lloyd1982least},
\changed{R1.11}{choosing 50 clusters in $\pmb{F}_{start}$ and $\pmb{F}_{end}$ and 5
clusters in $\pmb{F}_{dir}$. The cluster numbers were chosen manually to assure a large enough size of each cluster for generative model training, while also providing a sufficient resolution in the respective feature space. }.
\changed{R1.6}{As generative models, we implemented neural \gls{ODE} based
    normalizing flows \cite{dinh2016density}, \cite{chen2018neural}. We made
    this choice over alternative, more classical methods for density estimation,
    such as \acrlongpl{GMM} or variational autoencoders, since we found that 
the parameter distributions can be too complex to be reasonably 
captured by \acrshortpl{GMM}, and that the normalizing flow models were very robust over a big range of hyperparameters.}
We used a batch size of 128, 
1000 training iterations, and a network layout of two fully connected hidden layers with 64 neurons each. 
After training, the model accuracies are $91\pm 2.1 \%$ for $\pmb{F}_{start}$, $87\pm 3.3 \%$ 
for $\pmb{F}_{dir}$, and $59 \pm 4.1 \%$ for $\pmb{F}_{end}$ for the 3 DOF arm\changed{R1.2}{, $89\pm 3.2 \%$ for $\pmb{F}_{start}$, $73\pm 4.3 \%$ 
for $\pmb{F}_{dir}$, and $47 \pm 3.6 \%$ for $\pmb{F}_{end}$ for the mobile base, and $82\pm 4.2 \%$ for $\pmb{F}_{start}$, $55\pm 3.9 \%$ for $\pmb{F}_{dir}$, and $32 \pm 5.1 \%$ for $\pmb{F}_{end}$
for the combined system}.
\changed{R1.12}{The model accuracies were determined by sampling parameters from
the generative models, simulating the corresponding trajectories, and calculating the feature values. The accuracy is the percentage of samples that are assigned to the cluster the model was trained on.} Errors were calculated from 10 training runs per model with randomized initial weights.
Cluster entities for the robot in this example are generated and stored in the database for
the identification of \acrlongpl{CC}.

\paragraph*{Cognitive Core Creation}
Based on the defined \acrlongpl{BM}, the \acrlong{CC} is instantiated for all three robots
after exploration and clustering, and
inherits the label \emph{reach} (see also Figure \ref{fig:cc_ccm}).
For all systems, one cluster in the constrained \acrlong{FS} $\pmb{F}_{dir}$ was found 
with a centroid value of $\approx 0.9$.
Since the cluster fulfils the \acrlong{BM}'s minimum and maximum value constraints for this
\acrlong{FS}, the cluster is linked to the \acrlong{CC}.

\paragraph*{Cognitive Core Sampling}
One important aspect of the \acrlong{CC} is the ability to sample trajectories that 
conform to the underlying behavior model. In the current example, particle swarm optimization algorithms are used for the global optimization 
of the three feature models that are combined in the \emph{reach} \acrlong{CC}, while constraints and feature inputs are weighted equally. Due to the stochastic nature of this process, several sub-optimal solutions are generated during sampling that represent the corresponding behavior.

\changed{R1.2}{To evaluate the specificity of the 
    clustering procedure, we also define an alternative \emph{reach} \acrlong{CC} without the 
directness constraint. This \acrlong{CC} will generate behavior that reaches an end effector end position in any way possible. Figure \ref{fig:cc_reach_arm_base} shows samples from the constrained and the 
unconstrained \emph{reach} \acrlong{CC} for both the 3-DOF arm and the mobile base. The trajectories 
were obtained by simulating the parameters sampled from the \acrlongpl{CC}. All simulations and visualizations were performed in the MARS simulator \cite{LangoszMARS}, which is integrated in the \pname{Q-Rock} workflow. As expected, the constrained \emph{reach} \acrlong{CC} generates direct motion to the desired goal, whereas samples from the unconstrained one are more indirect. Note, however, that these indirect trajectories can be useful in 
different settings, e.g., when parts of the task space are obstructed. This example also illustrates that samples from the \acrlongpl{CC} do not exactly match the desired start and end position, and also show variability along the trajectory. Whereas this might seem counter-intuitive from a pure controlling standpoint, \acrlong{CC} sampling is deliberately probabilistic to allow generation of all possible behaviors that match the 
corresponding \acrlong{BM} description, and not one single optimal trajectory.} 

\changed{R1.2}{To demonstrate 
the applicability of our approach to more complex systems, we show samples from the \emph{reach} \acrlong{CC} 
of a combined system of 3-DOF arm and mobile base in Figure 
\ref{fig:reach_unconst_mob_base}. Here, combinations of motion of the mobile base and the 
attached arm are required to reach the target location of the end effector. As in the previous examples, the 
\emph{reach} \acrlong{CC} successfully generates direct motion towards the target location.}

\changed{R1.2}{Another way of generating more complex behavior for a system of already explored components is to use the previously generated \acrlongpl{CC} for each subsystem. Figure \ref{fig:cc_combined_reach} shows the parallel execution of both \emph{reach} \acrlongpl{CC} on each subsystem. The \acrlongpl{CC} of the arm and the mobile base are sampled individually to reach a target with the end effector of the arm that would lie outside of a fixed arm's task space.
As expected, the assumption that the task space trajectory of the combined system can be 
simply composed from the trajectories of the subsystems does not generally hold. However, 
the sampled motion can be a good first approximation of the actual behavior when the 
dynamical coupling between subsystems is not too strong.}

\paragraph*{Cognitive Core Annotation}
For the annotation step, the \acrlong{CC} is executed several times with
different variable inputs to 
$\pmb{F}_{start}$ and $\pmb{F}_{end}$. Videos of the performance of the
\acrlongpl{CC} are generated and shown to a user, who can confirm the
selected labels for the \acrlong{CC} or (re)assign labels. 
For this demonstration the user approves and sticks to the label \emph{reach} for
the \acrlong{CC}, which has been inherited from its \acrlong{BM}.

\subsection{Solving a Task}
\begin{figure}
    \centering
    \captionsetup{type=figure}
    \begin{subfigure}[t]{.45\textwidth}
        \includegraphics[width=\textwidth]{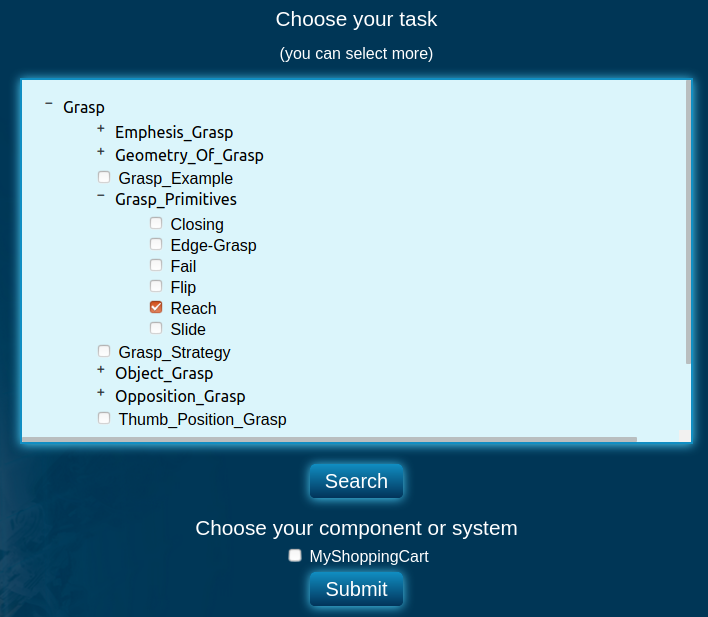}
        \caption{Selecting labels of the desired \acrlong{CC}}
        \label{fig:workflow:select_labels_for_query}
    \end{subfigure}
    \quad
    \begin{subfigure}[t]{.45\textwidth}
        \includegraphics[width=\textwidth]{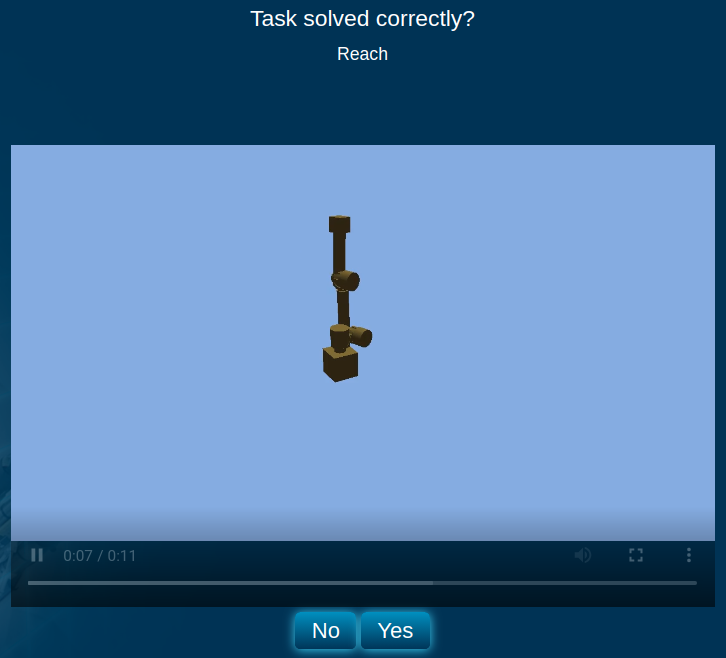}
        \caption{Validate the performance of the task}
        \label{fig:workflow:validate_performance}
    \end{subfigure}
    \caption{Query the database by matching against \acrlongpl{SA} of
    \acrlongpl{CC}}
    \label{fig:workflow:solve_my_task}
\end{figure}

\Acrlongpl{CC} are semantically annotated in order to provide a high-level description or
rather specification of their performance.
A user is not necessarily interested in designing new systems, but will
typically first search for available robots which can solve the task at
hand, or show similar performances.
For this use case, we designed the workflow named \emph{Solve My Task} (see
Figure~\ref{fig:workflow:solve_my_task}), where a user
selects a combination of semantic labels from the existing ontology and matches them
against existing \acrlongpl{CC} in the database.
Before the user has to make a final choice, the performance of each identified
\acrlong{CC} can be inspected through the previously rendered videos.
Here, the explored \emph{reach} \acrlong{CC} can successfully be retrieved
and visualized to the user.

\subsection{Solving a Mission}
\begin{figure}
    \centering
    \captionsetup{type=figure}
    \begin{subfigure}[t]{.45\textwidth}
        \includegraphics[width=\textwidth]{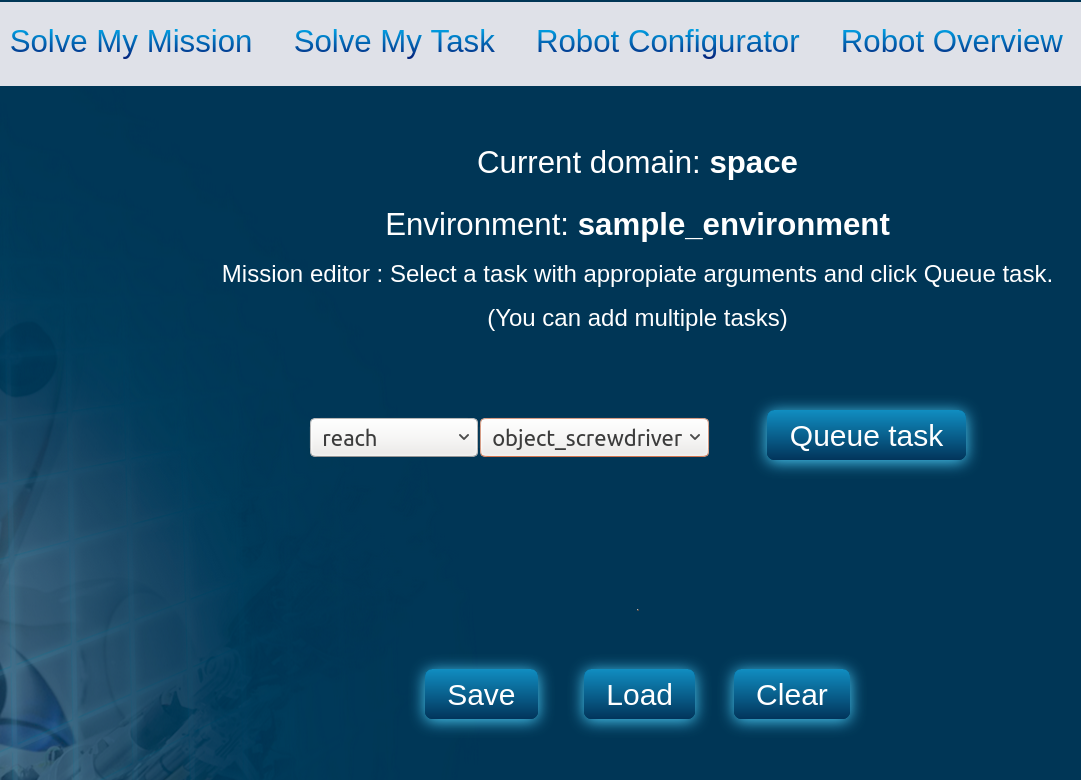}
        \caption{\changed{R1.26}{Design a mission with a web-editor}}
        \label{fig:workflow:design_mission}
    \end{subfigure}
    \quad
    \begin{subfigure}[t]{.45\textwidth}
        \includegraphics[width=\textwidth]{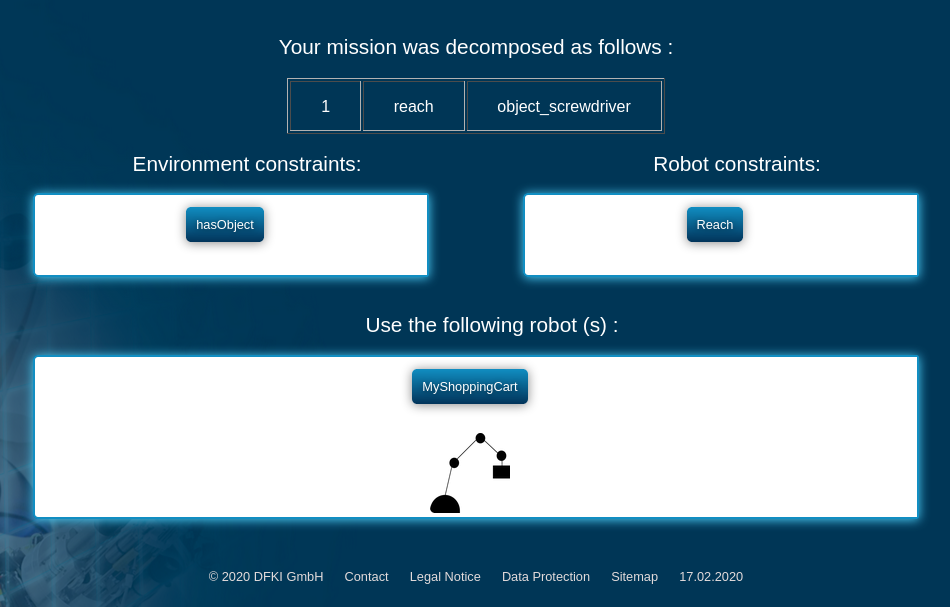}
        \caption{Suggest a robot, that can solve the mission, after matching the
            actions to suitable \acrlong{CC} - actions and \acrlongpl{CC} are
    linked via \acrlongpl{SA}}
        \label{fig:workflow:solved_mission}
    \end{subfigure}
    \caption{Solve a custom mission with robotic systems that are already in the database}
    \label{fig:workflow:solve_my_mission}
\end{figure}

The final use case, illustrated in Figure~\ref{fig:workflow:solve_my_mission}, deals with solving a user's application scenario, in the
following referred to as \emph{mission}. A mission can range from
a single robot action to a complex plan involving multiple actions that need to be sequenced.
A mission with sequential actions can be composed through our web interface,
based on a set of predefined - yet generic - actions: grasp, navigate,
perceive, pick, reach, release.
For this evaluation we select the \emph{reach} action, which maps to a requirement for
\acrlongpl{CC} with a \acrlong{SA} including the label "reach", so that as
an intermediate result the previously identified core can be picked.
The \acrlong{CC} is linked to the design of the 'NewShoppingCart', which can now
be considered a suitable robot system to perform the mission.
Therefore, this custom design is the final suggestion of the \pname{Q-Rock}
development cycle to solving this mission.

\changed{R1.15}{
\subsection{Summary}
The complementary use cases show the working of key elements of the \pname{Q-Rock}
development cycle.
Assemblies can be composed in a simple manner by reusing predefined
components from the \pname{Q-Rock} database. The assemblies are explored in simulation to
identify behavior and subsequent clustering of trajectories allows to extract \acrlongpl{CC}.
Semantic annotations bridge the gap between \acrlongpl{CC} and planning tasks, so that the
loop between the bottom-up system analysis and top-down requirement-based robot
design is closed.
Overall, although the use case for the \emph{reach} behavior is simple, this evaluation shows a fully integrated
workflow implementation. 
It validates the feasibility and working of the concept as first step towards automatically
developing complex robot behavior.
}

    \section{Discussion}
    \label{sec:discussion}
    In this paper, we introduced the formal concepts be-
hind Q-Rock and presented use cases to demonstrate
our approach to solving problems in a combination of
bottom-up and top-down solving. The use cases show
that the integrative workflow has been established, al-
though several challenges remain. In the following, we
will discuss key aspects of the workflow in more de-
tail.

\subsection{Exploration}
\label{disc:exploration}
The exploration framework has been successfully implemented and allows the automated data
generation for assemblies built with the robot configurator. The implementation
is designed in a modular way, so that minimal effort is needed to switch between
capability function models.

\changed{R1.1}{
In the current state the exploration is done in the configuration space of an
assembled system. This approach is not scaling well if the system complexity is
increasing, such as for systems of systems. Whereas for the combined system illustrated in Figure
\ref{fig:cc_combined_reach}, the full exploration approach is still feasible, it will break down
at some complexity level. To allow an application of the exploration approach to
these complex cases we have to make use of the knowledge already generated for the subsystems. For the simplest concept to reuse the subsystem capabilities, one would sample capabilities out of the subsystems clusters to generate capabilities for the combined system and thus form an exploration dataset. The concept development and evaluation is an ongoing process.}

After the training process, an implemented validation model can be queried for
the information whether a capability can be executed by the robot. However,
selecting a capability and then checking it is inconvenient for the clustering
process. For this reason, in the next step we are looking into inverting this
validity information, i.e.,
mapping all valid (and only the valid) capabilities into a continuous parameter space, on which the clustering can operate.

The most challenging problem for the exploration is to make these approaches scale
 with increasingly complex systems. In order to deal with this challenge we plan
 to apply more sophisticated search strategies. As the exploration is supposed
 to be task independent we intend to use intrinsic motivations \cite{aubret2019intrinsic} to explore the search space in a structured manner.

Another challenge is the exploration of perception capabilities which is
considered in the theoretical framework, but requires non-trivial environments to perform the exploration. Designing test environments to allow a mostly task independent exploration is one main challenge, besides the fact that for perception the capability space dimensionality is even higher compared to kinematic and dynamic exploration.

In parallel to the exploration approach, an introspection
into failure cases is envisaged via a hierarchical capability checking framework
which can (a) detect whether a given action is feasible on the robot and (b) pin-point the reasons of infeasibility. 
This problem is especially interesting for mechanisms with closed loops, e.g., parallel robots or serial-parallel hybrid robots~\cite{2020_Kumar_HybridRobots_Survey}.
We plan to exploit knowledge about the kinematic structure of the robot, its various physical properties, 
and analytical mappings between different spaces (actuator coordinates, generalized coordinates and full configuration space coordinates) by
using \gls{HyRoDyn} which is under active development in \pname{Q-Rock}.

\subsection{Classification \& Annotation}
Following exploration, capability clustering and the application of cognitive core and behavior model 
formalizations have also been successfully implemented,
while revealing interesting challenges. 

Clustering is an important step in the Classification \& Annotation workflow, since
it increases the granularity of the latent space on which the generative models are trained. 
Whereas the current implementation, using k-means clustering, has the advantage of being 
robust and well developed, a shortcoming is that if more finely spaced clusters are 
necessary for a specific behavior model, corresponding regions of the feature space have to 
be clustered again and cluster models trained anew. We are thus pursuing the
integration of more sophisticated generative model learning approaches into the
workflow that retain more accessible information about the underlying data distributions, such as arbitrary conditional flows \cite{li2020acflow}.

\Acrlongpl{BM} are also an important aspect in this part of \pname{Q-Rock}. Whereas only hand-crafted \acrlongpl{BM} and 
feature spaces have been tested to date, we aim at a more automated approach in the 
future. We actively research the application of variational autoencoders to world state trajectory data, and how well features found in this way can be semantically interpreted. Furthermore, we are working on automated extraction of behavior model definitions, 
both from human demonstration data and from modelling human evaluation functions.  

Although our goal is to increase the
level of automation in the future, we still see human labelling as a crucial backstop in the cycle to give meaning to the explored data and to
introduce steps for revision as part of the bottom-up path.
\changed{R2.2}{Recent work has also shown that pure automation based \acrshort{AI} approaches can be inferior to an interactive human in the loop in complex reasoning tasks \cite{holzinger2019interactive}. We thus see leveraging human 
semantic knowledge rather to be a feature of our approach than 
a shortcoming.}

An interesting theoretical problem is defining where the domain of cognitive
cores ends, and the domain of planning begins, i.e., up to which behavioral complexity level cognitive cores can be reasonably defined. The cognitive core formalism is purposefully flexible enough such that planning algorithms can be expressed, thus there is no clear limitation imposed on the formal side. A related challenge is parallel  execution of \acrlongpl{CC} of subsystems, such as demonstrated in Figure \ref{fig:cc_combined_reach}. Whether learning based techniques that are warm-started with initial guesses from individual \acrlong{CC} samples, higher level policy training on the established latent space of simple policies similar to \cite{florensa2017stochastic}, or more reasoning oriented approaches 
will prove more effective in our workflow is not finally resolved. An alternative solution is 
only using cognitive cores of a full system and using its subsystems' clusters for the 
exploration of the full system, as discussed in \ref{disc:exploration}.

\subsection{Reasoning}
\changed{R1.13}{To effectively exploit explored cognitive cores and behaviors, we suggest a
high-level planning based approach in this paper. 
Since the existing decomposition of problems is currently depending on the
planner's domain definition, we also provide an interface to create new
missions.
The current use case does not challenge this interface, since only very simple
tasks decompositions are required.
For increasingly complex problem descriptions, this mission description interface needs to remain not
only intuitive, but also sufficiently expressive.
Furthermore, this interface needs to be extensible through a growing vocabulary.
Another challenge remains for the application of a found solution and
exploitation of existing behaviors to address a user's high-level problem.
The framework provides first a mostly generic and robot-agnostic solution. This solution has
then to be mapped to the finally selected robot, i.e., this robot needs to execute the solution.
To achieve the latter, we have to run our \acrlong{CC} in robot-specific
contexts, e.g., exploiting existing robotic frameworks such as \acrshort{ROS}. 
Thus a grounding of generic solutions to selected environments and newly
designed robots has to be performed.
Semantic annotations are key elements for this mapping since they
provide the essential glue between reasoning and cognitive cores, where the
vocabulary is defined by our ontology.
}
We still need to enrich and revise the structure of this
ontology based on human feedback processes of annotating cognitive cores.
Hence, developing a sufficiently expressive semantic annotation
language remains a further challenge for the reasoning part.

The size of the database or rather number of components and options to combine
components in new systems leads to another, combinatorial challenge.
Here, we need to find effective heuristics in the context of the puzzler
development.

\subsection{Outlook}
From the view point of \pname{Q-Rock} as a whole, we see two major challenges arising
for future work. On the one hand, the system requires a rich database 
of annotated components, i.e., single parts or already simple robots, together with a design flow  to create new assemblies, in order to generate a significant added value for users.
A proposal for modelling components in such a design flow has been developed in the predecessor
project \pname{D-Rock} and is described in Section~\ref{sec:modelling_robot_composition}.
With this, \pname{Q-Rock} has to be thoroughly tested, such that new robotic devices are created and 
many \acrlongpl{CC} are built, which in turn fosters an enriched ontology to also interact with the user. During this process, it is important to evaluate the results of \pname{Q-Rock} in terms of completeness of found behaviors, as well as stability and robustness of
the underlying representations.

A related challenge is the introduction of \pname{Q-Rock} to a considerable number of users to start forming a community.
As a first step towards this goal, we intend to publish the parts of \pname{Q-Rock} open source, and a strategy for addressing the robotics
community is currently being formulated. The more users interact with \pname{Q-Rock}, and thereby also enrich the database, the more
individual users will benefit and the more versatile it will become.

Taking a further step back, the \pname{Q-Rock} system is part of a greater
development cycle in the \pname{X-Rock} project series. In D-Rock, the
groundwork was laid for simplified modelling of robot parts and construction of
robots based on well defined interfaces. In \pname{Q-Rock}, robots are enabled
to explore possible behaviors. In future projects, beyond the scope of currently
ongoing research, we plan to tackle questions regarding combinations of systems
and their respective cognitive cores, behavioral interactions between humans and
robots or groups of robots, and fine-tuning of behaviors for specific contexts.

    \section{Conclusions}
    \label{sec:conclusions}
    The fundamental idea behind the \pname{Q-Rock} approach is to integrate and extend existing methods in 
AI, both on the symbolic and the sub-symbolic level, and to implement a framework that assists users to solve
their intended task with an existing or novel robot. To achieve this, a central challenge is a unifying concept and theoretical 
\added{foundation} to (a) integrate all components in order to realize the
\pname{Q-Rock} \added{development} cycle (given in Figure~\ref{fig:qrock_cycle}), and (b), to have a clear definition of interfaces to extend the existing cycle or even replace single components with new 
approaches. \pname{Q-Rock} focuses on this
integration to set up a new way of designing complex robots with the help of AI and with the knowledge that previous designers
contributed to the knowledge base.

In this paper, we made an essential step by introducing the conceptual framework
as a basis \added{and integration platform} for all subsequent work. 
\added{Modularity is a key feature of our approach and allows embedding of alternative solutions for each stage. In the future, we expect competing or continuously improving implementations for each of
the stages of the development cycle.}
With the use cases presented in this paper, we already demonstrated the functional coupling
of all steps in \pname{Q-Rock}. In particular, the example shows that a model of annotated hardware can be 
used to successfully generate simple robotic capabilities (starting at E1 in the cycle), which can be successfully clustered
and annotated to generate a \acrlong{CC}. Hereby, a link is established from an exploration on the sub-symbolic level
to a representation containing a semantic label, such that semantic input from a user can be made on which reasoning is
performed. \pname{Q-Rock} is unique in this way: goal-agnostic capabilities are cast into a broader semantic framework, and sub-symbolic and symbolic levels in AI are integrated.

    \section{Appendix}
    \label{sec:appendix}

\printglossary[type=\acronymtype,title={Abbreviations}, nonumberlist]

    \section{Declarations}

    \begin{authordeclaration}{Ethical Approval and Consent}
        Not applicable.
    \end{authordeclaration}

    \begin{authordeclaration}{Consent for publication}
        All authors gave their consent to publish this version of the manuscript.
    \end{authordeclaration}

    \begin{authordeclaration}{Availability of data and materials}
    The datasets generated and analysed for the presented use case scenario are
    not publicly available.
    \end{authordeclaration}

    \begin{authordeclaration}{Competing interests}
    The authors declare that they have no competing interests.
\end{authordeclaration}

    \begin{authordeclaration}{Funding}
            This research  and  development  project  is  funded  by  the
            German Federal Ministry of Education and Research under grant
            agreement (FKZ 01IW18003).
    \end{authordeclaration}

    \begin{authordeclaration}{Authors' Contributions}
            T.R. wrote the manuscript, developed software and theory. D.H. wrote the manuscript, developed software and theory. H.W. supervised the project, provided text for sections \ref{sec:introduction} and \ref{sec:discussion}. F.W. developed software, theory and provided text for sections \ref{sec:exploration} and \ref{sec:discussion}. M.S. developed software, theory and provided text for section \ref{sec:modelling_robot_composition}, developed theory for \ref{sec:reason:bottom_up}. O.L. developed software, theory and provided text for section \ref{sec:reasoning}. M.L. developed software and theory and provided text for sections \ref{sec:sota}, \ref{sec:exploration} and \ref{sec:discussion}. S.K. provided text for section \ref{sec:discussion}. S.S. supervised the project, provided text for sections \ref{sec:introduction} and \ref{sec:discussion}. F.K. conceived of the original idea, supervised the project, provided text for sections \ref{sec:introduction} and \ref{sec:discussion}.
    \end{authordeclaration}

    \begin{authordeclaration}{Acknowledgements}
            We thank all members of the Q-Rock development team and the
            internal reviewers for their valuable feedback on preliminary
            versions of this paper.
    \end{authordeclaration}

    \begin{authordeclaration}{Authors' information}
        (Optional - no information provided)
    \end{authordeclaration}



    \bibliographystyle{styles/IEEEtran}
    \balance
    \bibliography{references}
    
\end{document}